\documentclass[sigconf]{acmart}

\usepackage{graphicx}
\usepackage{amsmath}
\usepackage{subfigure}
\usepackage{amsfonts}
\usepackage{stfloats}
\usepackage{multirow}

\AtBeginDocument{%
  }

\copyrightyear{2023}
\acmYear{2023}
\setcopyright{acmlicensed}
\acmConference[MM '23] {Proceedings of the 31st ACM International Conference on Multimedia}{October 29--November 3, 2023}{Ottawa, ON, Canada.}
\acmBooktitle{Proceedings of the 31st ACM International Conference on Multimedia (MM '23), October 29--November 3, 2023, Ottawa, ON, Canada}
\acmPrice{15.00}
\acmISBN{979-8-4007-0108-5/23/10}
\acmDOI{10.1145/3581783.3611792}

\graphicspath{{fig1/},{fig2/},{fig3/},{fig4/},{fig5/},{fig6/}}
\begin{document}

\title{Calibration-based Dual Prototypical Contrastive Learning \\Approach for Domain Generalization Semantic Segmentation}


\author{Muxin Liao}
\thanks{Muxin Liao and Shishun Tian contributed equally to this work.}
\orcid{0000-0002-8461-1946}
\affiliation{%
	\institution{Guangdong Key Laboratory of Intelligent Information Processing}
	\institution{College of Electronics and Information Engineering}
	\institution{Shenzhen University}
	\city{Shenzhen}
	\country{China}
}
\email{liaomuxin2020@email.szu.edu.cn}

\author{Shishun Tian}
\affiliation{%
	\institution{Guangdong Key Laboratory of Intelligent Information Processing}
	\institution{College of Electronics and Information Engineering}
	\institution{Shenzhen University}
	\city{Shenzhen}
	\country{China}}
\email{stian@szu.edu.cn}

\author{Yuhang Zhang}
\affiliation{%
	\institution{Guangdong Key Laboratory of Intelligent Information Processing}
	\institution{College of Electronics and Information Engineering}
	\institution{Shenzhen University}
	\city{Shenzhen}
	\country{China}
}
\email{zhangyuhang2019@email.szu.edu.cn}

\author{Guoguang Hua}
\affiliation{%
	\institution{Guangdong Key Laboratory of Intelligent Information Processing}
	\institution{College of Electronics and Information Engineering}
	\institution{Shenzhen University}
	\city{Shenzhen}
	\country{China}}
\email{huaguoguang2021@email.szu.edu.cn}

\author{Wenbin Zou}
\authornote{Corresponding author.}
\affiliation{%
	\institution{Guangdong Key Laboratory of Intelligent Information Processing}
	\institution{College of Electronics and Information Engineering}
	\institution{Shenzhen University}
	\city{Shenzhen}
	\country{China}}
\email{wzou@szu.edu.cn}

\author{Xia Li}
\affiliation{%
	\institution{Guangdong Key Laboratory of Intelligent Information Processing}
	\institution{College of Electronics and Information Engineering}
	\institution{Shenzhen University}
	\city{Shenzhen}
	\country{China}}
\email{lixia@szu.edu.cn}

\renewcommand{\shortauthors}{Muxin Liao et al.}


\begin{abstract}
  Prototypical contrastive learning (PCL) has been widely used to learn class-wise domain-invariant features recently. These methods are based on the assumption that the prototypes, which are represented as the central value of the same class in a certain domain, are domain-invariant. Since the prototypes of different domains have discrepancies as well, the class-wise domain-invariant features learned from the source domain by PCL need to be aligned with the prototypes of other domains simultaneously. However, the prototypes of the same class in different domains may be different while the prototypes of different classes may be similar, which may affect the learning of class-wise domain-invariant features. Based on these observations, a calibration-based dual prototypical contrastive learning (CDPCL) approach is proposed to reduce the domain discrepancy between the learned class-wise features and the prototypes of different domains for domain generalization semantic segmentation. It contains an uncertainty-guided PCL (UPCL) and a hard-weighted PCL (HPCL). Since the domain discrepancies of the prototypes of different classes may be different, we propose an uncertainty probability matrix to represent the domain discrepancies of the prototypes of all the classes. The UPCL estimates the uncertainty probability matrix to calibrate the weights of the prototypes during the PCL. Moreover, considering that the prototypes of different classes may be similar in some circumstances, which means these prototypes are hard-aligned, the HPCL is proposed to generate a hard-weighted matrix to calibrate the weights of the hard-aligned prototypes during the PCL. Extensive experiments demonstrate that our approach achieves superior performance over current approaches on domain generalization semantic segmentation tasks.
\end{abstract}




\maketitle

\begin{figure*}[!t]
	\centering
	\begin{center}
		\includegraphics[height=4.5cm,width=13cm]{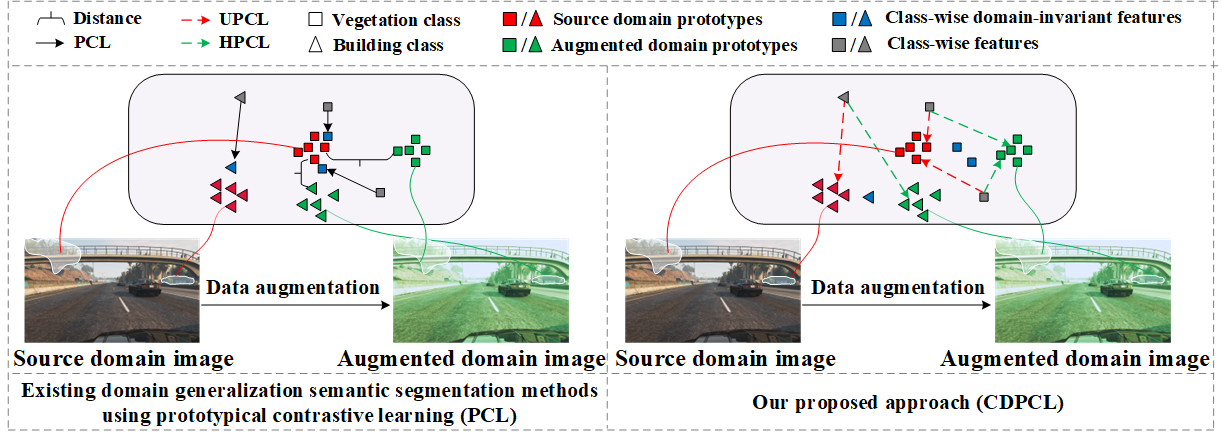}
	\end{center}
	\caption{The illustration of the domain discrepancy between the learned class-wise features and the prototypes of different domains. The augmented domain is generated from the source domain by using data augmentation.}
	\label{fig:intro}
\end{figure*}

\section{Introduction}
\label{sec:intro}
Semantic segmentation plays an important role in multiple real-world applications, such as autonomous driving \cite{he2021fast, ma2022rethinking}, environment understanding \cite{hua2023multiple, zhang2022self, li2022cross}, and medical diagnosing \cite{ye2022alleviating, zhang2022probabilistic}. With the rapid development of deep neural networks, supervised semantic segmentation methods \cite{chen2017deeplab, chen2018encoder, wei2022sginet} have achieved remarkable progress on the independent and identically distributed assumption. However, the performance of these methods dramatically degrades when they are applied to target domain data, due to the domain discrepancy problem between the training data (source domain) and the testing data (target domain). Collecting abundant target domain images and annotating each pixel for images to retrain the model is one possible solution that is expensive and time-consuming. Thus, unsupervised domain adaptation semantic segmentation (UDASS) methods \cite{zhang2023hybrid, liao2022exploring, zou2023dual} are proposed to address the domain discrepancy problem.

The key to UDASS methods is to learn domain-invariant features from the labeled source domain and the unlabeled target domain. Although UDASS methods achieve significant performance, they still have limitations. First, UDASS methods could perform well on the target domain but their performances sharply degrade when evaluating out-of-distribution scenes. Second, collecting sufficiently various out-of-distribution data that covers all scenes is impractical and even impossible. To address these limitations, domain generalization methods \cite{wang2022domain,wang2022feature} are proposed.

Domain generalization is a more practical and challenging setting than domain adaptation since any target domain data is not accessed during the training process. Thus, the key to domain generalization is to learn domain-invariant features from single or multiple labeled source domains. Recent methods \cite{lu2022bidirectional, jiang2022prototypical, chen2022compound} have been proposed to learn class-wise domain-invariant features from the prototypical contrastive learning (PCL). These methods are based on the assumption that the prototypes, which are represented as the central value of the same class from different domains, are domain-invariant. Since the prototypes of different domains have discrepancies as well \cite{lee2022bi}, the class-wise domain-invariant features learned from the source domain by PCL need to align with the prototypes of the other domains simultaneously. However, the prototypes of the same class in different domains may be different while the prototypes of different classes may be similar \cite{wang2020learning}, which may affect the learning of class-wise domain-invariant features. As shown in the left of Figure \ref{fig:intro}, the prototypes of the ``vegetation'' class between the source and augmented domains have discrepancy, where the augmented domain is generated from the source domain by using data augmentation. Thus, there is still domain discrepancy between the class-wise domain-invariant features learned from the source domain by PCL and the prototypes of the augmented domain. Moreover, since the prototypes of the ``building'' class in the source domain and the prototypes of the ``vegetation'' class in the augmented domain may be similar, the domain-invariant features of the ``vegetation'' class learned from the source domain may be close to the ``building'' class prototypes of the augmented domain. Based on these observations, a calibration-based dual prototypical contrastive learning (CDPCL) approach is proposed to reduce the domain discrepancy between the learned class-wise features and the prototypes of different domains for domain generalization semantic segmentation.

The CDPCL approach contains an uncertainty-guided PCL (UPCL) and a hard-weighted PCL (HPCL). Since the domain discrepancies of the prototypes of different classes may be different \cite{lee2022bi}, we propose an uncertainty probability matrix to represent the domain discrepancies of the prototypes of all the classes. The UPCL estimates the uncertainty probability matrix to calibrate the weights of the prototypes during the PCL. In the uncertainty probability matrix, a small probability means a big difference between the prototypes of different domains. Thus, the uncertainty probability matrix can be set as the weight matrix for the calibration of prototypes. Moreover, considering that the prototypes of different classes may be similar in some circumstances \cite{wang2020learning}, which means these prototypes are hard-aligned \cite{wang2023bp}, the HPCL is proposed to generate a hard-weighted matrix by computing the similarity between the prototypes of different classes for calibrating the weights of the hard-aligned prototypes during the PCL. As shown in the right of Figure \ref{fig:intro}, the class-wise domain-invariant features learned from our proposed approach are close to the corresponding prototypes of the source and augmented domains and far from the prototypes of different classes. Extensive experiments demonstrate that our approach achieves superior performance over current approaches on domain generalization semantic segmentation. The contributions are summarized as follows:
\begin{itemize}
	\item[1.] This paper proposes a calibration-based dual prototypical contrastive learning approach to reduce the domain discrepancy between the learned class-wise features and the prototypes of different domains for domain generalization semantic segmentation.
	\item[2.] We propose an uncertainty-guided prototypical contrastive learning to estimate an uncertainty probability matrix for calibrating the weights of the source domain prototypes during the PCL. 
	\item[3.] We propose a hard-weighted prototypical contrastive learning to generate a hard-weighted matrix for calibrating the weights of the augmented domain prototypes during the PCL. 
	\item[4.] The proposed approach achieves superior performance against the state-of-the-art methods on multiple challenging tasks for domain generalization semantic segmentation.
\end{itemize} 

\section{Related work}
\label{sec:rel_work}

\subsection{Domain Generalization Methods for Semantic Segmentation}
Existing domain generalization semantic segmentation (DGSS) methods are mainly divided into three types: domain randomization methods \cite{zhao2022style, su2022consistency}, normalization and whitening methods \cite{choi2021robustnet, peng2022semantic}, and meta-learning-based method \cite{kim2022pin}.

Domain randomization methods randomly generate different styles in the input space \cite{yue2019domain, peng2021global, zhao2022style, tjio2022adversarial} or the feature space \cite{huang2021fsdr, lee2022wildnet, su2022consistency} to train a style-insensitive model. For domain randomization methods used in the input space, Yue et al. \cite{yue2019domain} propose to randomize the style of the source domain images by using CycleGAN \cite{zhu2017unpaired} to learn style-insensitive features. Different from \cite{yue2019domain}, these methods \cite{peng2021global, zhao2022style, tjio2022adversarial} propose to use a hallucinatory style strategy for randomizing the style of the source domain images. For domain randomization methods used in the feature space, Huang et al. \cite{huang2021fsdr} propose to transform the features from the spatial domain to the frequency domain and then randomize the style in the frequency domain. Except for randomizing the style of the source domain images, Lee et al. \cite{lee2022wildnet} propose to randomize the content borrowed from the ImageNet for learning class-discriminant features. Different from \cite{huang2021fsdr, lee2022wildnet} which perform stylization on coarse-grained image-level features, Su et al. \cite{su2022consistency} propose a class-aware style variation method to generate fine-grained class-aware stylized images for learning class-level domain-invariant features.

Normalization and whitening methods \cite{pan2018two, pan2019switchable, choi2021robustnet, xu2022dirl} utilize different style normalization strategies, such as instance normalization \cite{pan2018two} or instance selective whitening \cite{choi2021robustnet}, for learning domain-invariant features. However, normalizing or whitening domain-specific features inevitably remove some task-relevant discriminative information, which may affect the performance. To address this issue, Peng et al. \cite{peng2022semantic} propose semantic-aware normalization and semantic-aware whitening to encourage both intra-category compactness and inter-category separability for enhancing the discriminability of networks. Moreover, based on the meta-learning framework, Kim et al. \cite{kim2022pin} propose to use an externally settled memory that contains the prototype information of classes for guiding the learning of domain-invariant features.

\subsection{Prototypical Contrastive Learning}
Prototypical contrastive learning have been utilized in many domain adaptation visual tasks, such as image classification \cite{chen2019progressive}, object detection \cite{zhang2021rpn, yu2022mttrans}, image semantic segmentation \cite{lu2022bidirectional, jiang2022prototypical}, and LiDAR point clouds semantic segmentation \cite{yuan2023prototype}. These methods claim that the prototypes are beneficial for learning class-wise domain-invariant features since they assume that the prototypes, which are represented as the central value of the same class from different domains, are domain-invariant. Since the prototypes of different domains have discrepancies as well \cite{lee2022bi}, the class-wise domain-invariant features learned from the source domain by PCL need to align with the prototypes of the different domains simultaneously. In the domain adaptation setting, Lee et al. \cite{lee2022bi} propose a calibration method to compensate for domain discrepancy of prototypes by the distance between the prototypes of the source and target domains. However, the target domain data is not accessed during the training process in the domain generalization setting, which means the prototypes of the target domain are agnostic. Moreover, the prototypes of the same class in different domains may be different while the prototypes of different classes may be similar \cite{wang2020learning}, which may affect the learning of class-wise domain-invariant features. 

To address these issues, a calibration-based dual prototypical contrastive learning approach is proposed to reduce the domain discrepancy between the learned class-wise features and the prototypes of different domains for domain generalization semantic segmentation.

\subsection{Uncertainty-guided Methods for Domain Generalization}
In this section, we discuss recent uncertainty-guided methods for domain generalization. Cai et al. \cite{cai2020generalizing} propose a Bayesian CNN-based framework to estimate the model uncertainty for guiding an iterative self-training method. Qiao et al. \cite{qiao2021uncertainty} propose a Bayesian meta-learning framework that aims to increase the capacity of input and label spaces from the source domain by using an uncertainty-guided augmentation strategy. Peng et al. \cite{peng2022out} propose an uncertainty-guided domain generalization method to quantify the generalization uncertainty which is used to guide the feature and label augmentation strategies. Xiao et al. \cite{xiao2021bit} propose a probabilistic framework via variational Bayesian inference to learn domain-invariant features by incorporating uncertainty into neural network weights.

Different from these uncertainty-guided domain generalization methods, we propose an uncertainty-guided prototypical contrastive learning to estimate an uncertainty probability matrix for calibrating the weights of the source domain prototypes during the prototypical contrastive learning.

\begin{figure*}[!t]
	\centering
	\includegraphics[width=1\textwidth,height=4.5cm]{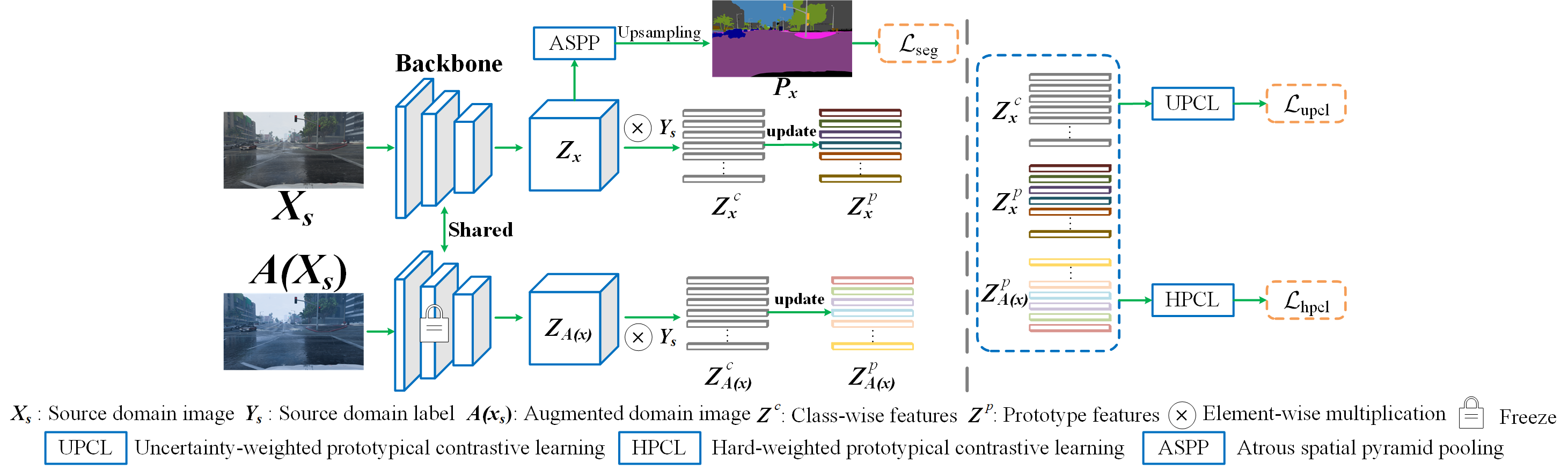}
	\caption{The proposed calibration-based dual prototypical contrastive learning (CDPCL) approach for domain generalization semantic segmentation.}
	\label{fig:framework}
\end{figure*}

\section{Approach}

\subsection{Problem Statement and Overview}
Given the labeled dataset as the source domain $S = \{x^s, y^s\}$, the goal of domain generalization is to train a model on the source domain which performs well on unseen domains. Recent methods \cite{lu2022bidirectional, jiang2022prototypical, chen2022compound} aim to learn class-wise domain-invariant features for improving the generalization ability by using prototypical contrastive learning which is denoted as follows:
\begin{equation}
\mathcal{L}_{pcl} = -\sum_{i=1}^{C}y_s^i\log \frac{exp(Z_x^{p_i}\cdot Z_x^{c_i}/\tau)}{\sum_{i\neq k}^{C} exp(Z_x^{p_k}\cdot Z_x^{c_i}/\tau)}
\label{contrast_sim_compute}
\end{equation}
where $Z_x^{p_i}$ is the prototype of the $i$th class. $Z_x^{c_i}$ is the features of the $i$th class. $\tau$ is the temperature parameter. The class-wise domain-invariant features are learned by minimizing the contrastive learning loss $\mathcal{L}_{pcl}$. The minimized loss means the distance between the $Z_x^{p_i}$ and $Z_x^{c_i}$ which represent the same class is decreased and the distance between the $Z_x^{p_k}$ and $Z_x^{c_i}$ which represent different classes is increased. 

Since the prototypes of different domains may have discrepancies as well \cite{lee2022bi}, the class-wise domain-invariant features learned from the source domain by PCL need to align with the prototypes of the different domains simultaneously. However, the prototypes of the same class in different domains may be different while the prototypes of different classes may be similar \cite{wang2020learning}, which may affect the learning of class-wise domain-invariant features. Based on these observations, a calibration-based dual prototypical contrastive learning (CDPCL) approach is proposed to reduce the domain discrepancy between the learned class-wise features and the prototypes of different domains for domain generalization semantic segmentation. The proposed approach is illustrated in Figure \ref{fig:framework}. 

The approach is divided into three parts: a semantic segmentation network, an uncertainty-weighted prototypical contrastive learning (UPCL), and a hard-weighted prototypical contrastive learning (HPCL). Specifically, the semantic segmentation network $F$ is utilized to extract the features $Z_x$ from the source domain images $X_s$. For one thing, the cross-entropy loss $\mathcal{L}_{seg}$ between the predicted segmentation map $P_x$ and the ground truth $Y_s$ is utilized to optimize the feature $Z_x$. The cross-entropy loss $\mathcal{L}_{seg}$ is denoted as follows:
\begin{equation}
\mathcal{L}_{seg}(P_x,Y_s)=-Y_s \cdot \log(P_x)
\label{loss:seg}
\end{equation}
where the predicted segmentation map $P_x$ is obtained by upsampling the feature $Z_{x}$ after putting into the Atrous Spatial Pyramid Pooling module. The $Y_s$ is the ground truth.

\begin{figure*}[!t]
	\centering
	\includegraphics[width=1\textwidth,height=3.5cm]{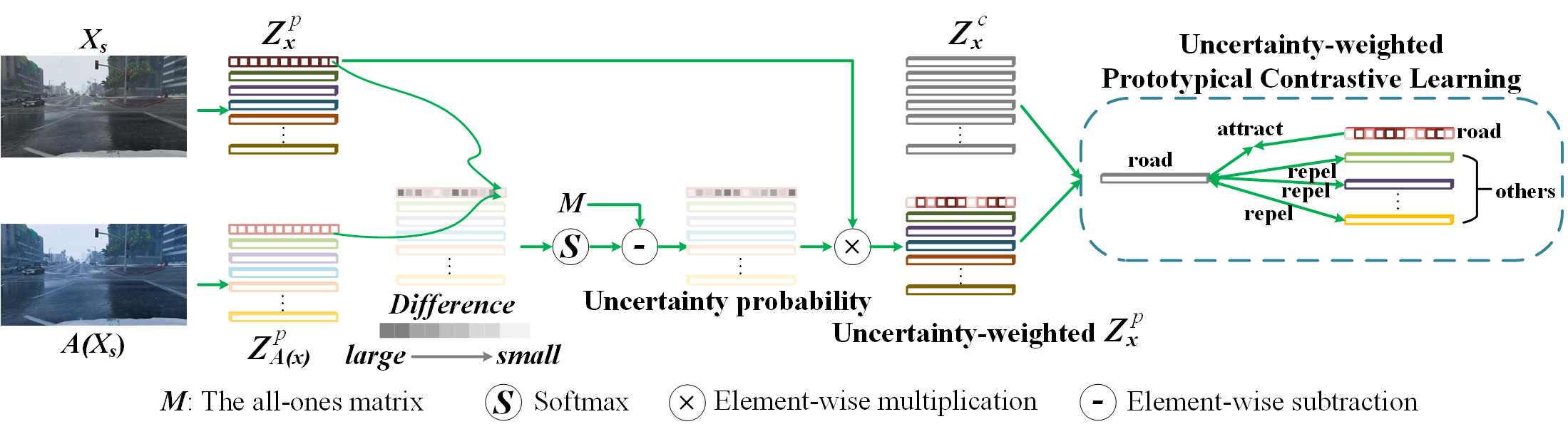}
	\caption{The illustration of the proposed uncertainty-weighted prototypical contrastive learning.}
	\label{fig:upcl}
\end{figure*}

For another, the features $Z_x$ are leveraged to generate the prototypes $Z_x^p$ for each class of the source domain. To reduce the domain discrepancy between the class-wise features $Z_x^c$ and the prototypes of the unseen domains, we first construct the augmented prototypes $Z_{A(x)}^p$ from the augmented domains. The augmented domains generated from the source domain by using data augmentation are considered as other unseen domains. Compared with the source domain images, the augmented domain images $A(X_s)$ have the same content but different styles. Then, the augmented images $A(X_s)$ are fed into the backbone to extract the augmented features $Z_{A(x)}$, where the weights of the backbone are frozen. Then, the augmented features $Z_{A(x)}$ are utilized to generate the augmented prototypes $Z_{A(x)}^p$. Finally, the UPCL and HPCL are utilized to reduce the domain discrepancy between the class-wise features $Z_x^c$ and the two prototypes $Z_x^p$ and $Z_{A(x)}^p$. In the following subsections, we sequentially introduce the uncertainty-weighted prototypical contrastive learning, the hard-weighted prototypical contrastive learning, and our total training loss.

\subsection{The Uncertainty-weighted Prototypical Contrastive Learning (UPCL)} 
Since the domain discrepancies of the prototypes of different classes may be different, we propose an uncertainty probability matrix to represent the domain discrepancies of the prototypes of all the classes. The UPCL estimates the uncertainty probability matrix to calibrate the weights of prototypes during the PCL for better learning class-wise domain-invariant features. 

Specifically, the difference matrix $D$ is first obtained by using the Manhattan Distance to compute the domain discrepancy between the prototypes $Z_x^p$ and the augmented prototypes $Z_{A(x)}^p$, which is denoted as follows:
\begin{equation}
D(Z_x^p,Z_{A(x)}^p)=||Z_x^p - Z_{A(x)}^p||_1
\label{eqa:diff}
\end{equation}
where $Z_x^p$ and $Z_{A(x)}^p \in \mathcal{R}^{C\times N}$. The $C$ and $N$ respectively denote the number of classes and the number of features. A big value of $D$ indicates a big difference between the prototypes of the two domains. The prototypes $Z_x^p$ and the augmented prototypes $Z_{A(x)}^p$ are updated in every iteration during the training process, which are denoted as follows:
\begin{equation}
Z_x^p = m_p Z_x^p + (1-m_p) Z_x^c
\label{eqa:proto_update}
\end{equation}
\begin{equation}
Z_{A(x)}^p = m_a Z_{A(x)}^p + (1-m_a) Z_{A(x)}^c
\label{eqa:aug_proto_update}
\end{equation}
where $m_p$ and $m_a$ are the trade-off parameters. $Z_x^c$ and $Z_{A(x)}^c$ are the class-wise features respectively obtained from the source domain images and the augmented domain images.
 
Second, to avoid the class-wise features $Z_x^c$ being aligned with the prototypes with big difference during the PCL, an uncertainty probability matrix is generated to calibrate the weights of these prototypes. The calibration of the prototypes assigns a small weight to the prototypes which have a big difference. To achieve this goals, the uncertainty probability matrix $U$ is obtained by subtracting the probability of the difference matrix $D$ from an all-ones matrix $M$, which is denoted as follows:
\begin{equation}
U=M - Softmax(D)
\label{eqa:upm}
\end{equation}
where $Softmax(D)$ denotes that the probability of the difference matrix $D$ is computed by using the Softmax in the dimensionality of class. In the uncertainty probability matrix $U$, a small probability means a big difference between the prototypes of the source and augmented domains. Then, the uncertainty probability matrix $U$ is leveraged to calibrate the weights of the prototypes $Z_x^p$ by using element-wise multiplication. Finally, the uncertainty-weighted prototypes are utilized for the UPCL which is denoted as follows by rewriting Eq. (\ref{contrast_sim_compute}):
\begin{equation}
\mathcal{L}_{upcl} = -\sum_{i=1}^{C}y_s^i\log \frac{exp((Z_x^{p_i} \times U_i)\cdot Z_x^{c_i}/\tau_u)}{\sum_{i\neq k}^{C} exp((Z_x^{p_k}\times U_k)\cdot Z_x^{c_i}/\tau_u)}
\label{loss:upcl}
\end{equation}
where $\times$ means element-wise multiplication. $\tau_u$ is the temperature parameter. In particular, the uncertainty probability matrix $U$ is updated when the augmented prototypes $Z_{A(x)}^p$ is updated, which is denoted as follows:
\begin{equation}
U = m_u U + (1-m_u) U_c
\label{eqa:u_update}
\end{equation}
where $m_u$ is a trade-off parameter. $U_c$ is the current uncertainty probability matrix.

\subsection{The Hard-weighted Prototypical Contrastive Learning (HPCL)}
Since the prototypes of different classes may be similar in some circumstances \cite{wang2020learning}, which means these prototypes are hard-aligned \cite{wang2023bp}, the HPCL is proposed to generate a hard-weighted matrix to calibrate the weights of the hard-aligned prototypes during the PCL for better learning class-wise domain-invariant features. 

\begin{figure}[htb]
	\centering
	\includegraphics[width=0.45\textwidth,height=4.5cm]{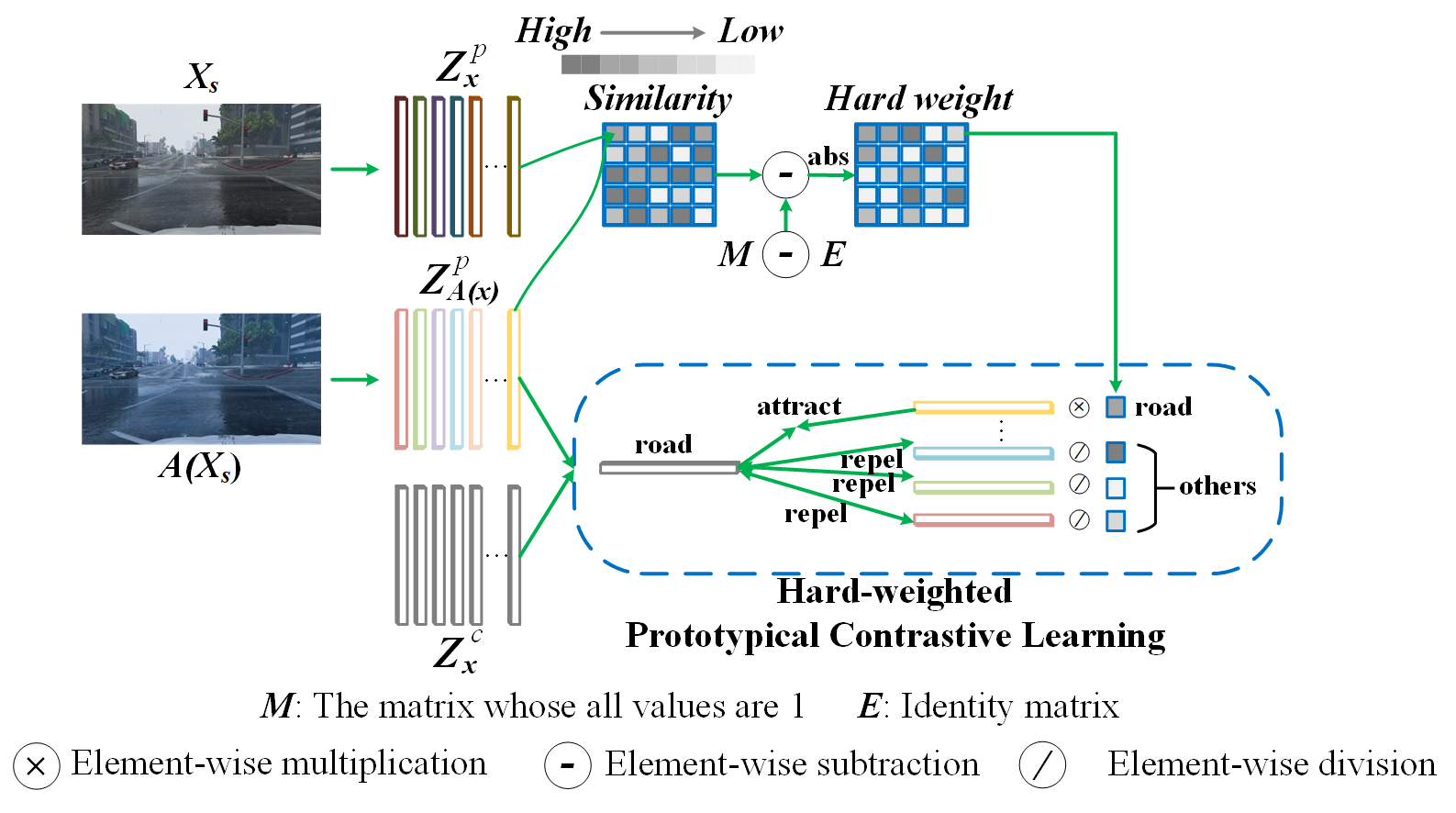}
	\caption{The illustration of the proposed hard-weighted prototypical contrastive learning.}
	\label{fig:hpcl}
\end{figure}

\begin{table*}[!t]
	\centering
	{
		\caption{Performance comparison in terms of mIoU (\%) between domain generalization methods in three architectures of the ResNet-50 \cite{he2016deep}, the ShuffleNetV2 \cite{ma2018shufflenet}, and MobileNetV2 \cite{sandler2018mobilenetv2} backbones. The best results are marked in bold and the second best results are underlined. $\dagger$ denotes that the performance is obtained by our reproduction of the respective method.} 
		\label{tab:all}
		\renewcommand\tabcolsep{1.0pt}
		\begin{tabular}{lcccccccccccccccccccccccccc}
			\hline
			\multirow{2}{*}{Methods}& 
			\multirow{2}{*}{Backbone}& 
			\multirow{2}{*}{Mean}& 
			\multicolumn{4}{c}{Train on G}&  &
			\multicolumn{4}{c}{Train on S}& & 
			\multicolumn{4}{c}{Train on C}&  &
			\multicolumn{4}{c}{Train on B}& &
			\multicolumn{4}{c}{Train on M} \\
			\cline{4-7}\cline{9-12}\cline{14-17}\cline{19-22}\cline{24-27}
			&&&  $\rightarrow$C&$\rightarrow$B&$\rightarrow$M&$\rightarrow$S&& $\rightarrow$C&$\rightarrow$B&$\rightarrow$M&$\rightarrow$G&& $\rightarrow$B&$\rightarrow$M&$\rightarrow$G&$\rightarrow$S&& $\rightarrow$G&$\rightarrow$S&$\rightarrow$C&$\rightarrow$M&& $\rightarrow$G&$\rightarrow$S&$\rightarrow$C&$\rightarrow$B \\
			\hline
			
			Deeplabv3+ \cite{chen2018encoder} & \multirow{12}{*}{ResNet-50}&29.9 &29.3&25.7&28.3&26.2&& 23.2&24.5&21.8&26.3&&45.2 &51.5 &42.6 &24.3 && 26.1 &21.7 &39.0 &23.9 && 25.5 &23.4 &36.8 &26.4 \\
			
			IBN \cite{pan2018two}& &34.2 &33.9&32.3&37.8&27.9&& 32.0&30.6&32.2&26.9&& 48.6&57.0&45.1&26.1&& 29.0&25.4&41.1&26.6&& 30.7&27.0&42.8&31.0 \\
			
			SW \cite{pan2019switchable}& &32.3 &29.9&27.5&29.7&27.6&& 28.2&27.1&26.3&26.5&& 48.5&55.8&44.9&26.1&& 27.7&25.4&40.9&25.8&& 28.5&27.4&40.7&30.5 \\
			
			DRPC \cite{yue2019domain}& &35.8 &37.4&32.1&34.1&28.1&& 35.7&31.5&32.7&28.8&& 49.9&56.3&45.6&26.6&& 33.2&29.8&41.3&31.9&& 33.0&29.6&46.2&32.9 \\
			
			GTR \cite{peng2021global}& &36.1 &37.5&33.8&34.5&28.2&& 36.8&32.0&32.9&28.0&& 50.8&57.2&45.8&26.5&& 33.2&30.6&42.6&30.7&& 32.9&30.3&45.8&32.6 \\
			
			ISW \cite{choi2021robustnet}& &36.4 &36.6&35.2&40.3&28.3&& 35.8&31.6&30.8&27.7&& 50.7&58.6&45.0&26.2&& 32.7&30.5&43.5&31.6&& 33.4&30.2&46.4&32.6 \\
			
			SAN \cite{peng2022semantic}& &38.5 &39.8&37.3&41.9&30.8&& 38.9&\underline{35.2} &34.5&29.2&& \textbf{53.0}&\underline{59.8}&\underline{47.3}&28.3&& 34.8&\underline{31.8} &44.9&33.2&& 34.0&\underline{31.6} &48.7&34.6 \\
			
			PinMem $\dagger$ \cite{kim2022pin} & &41.0 &41.2 &35.2 &39.4 &28.9 && 38.2 &32.3 &33.9 &32.1 &&50.6 &57.9 &45.1 &29.4 && 42.4 &29.1 &54.8 &51.0 && 44.1 &30.8 &55.9 &47.6 \\
			
			WildNet $\dagger$ \cite{lee2022wildnet} & &\underline{42.6} &\textbf{44.6} &38.4 &\textbf{46.1} &\underline{31.3} && 38.4&33.5&32.8&\underline{34.9} && 50.9 &58.8 &47.0 &28.0 && \underline{45.1} &30.2 &55.7 &\underline{54.1} && \underline{46.1} &30.2 &\underline{57.1} &\underline{49.2} \\
			
			SHADE $\dagger$ \cite{zhao2022style} & &42.5 &\underline{43.5} &\textbf{40.3} &\underline{43.0} &31.2 && \underline{39.6} &29.2 &\underline{34.7} &34.8 &&51.5 &58.7 &\textbf{48.2} &\underline{30.8} && 43.5 &31.1 &\underline{56.2} &53.1 && 45.3 &31.1 &56.2 &48.5 \\
			
			Ours& & \textbf{45.2}& 42.2& \underline{39.5}& 42.4& \textbf{33.0}&&  \textbf{41.2}& \textbf{35.4}& \textbf{35.5}& \textbf{36.6}&&  \underline{52.2} &  \textbf{60.7}&  46.8& \textbf{31.9}&&  \textbf{49.6}& \textbf{34.2}& \textbf{58.1}& \textbf{57.3}&&  \textbf{51.3}& \textbf{36.2}& \textbf{65.0}& \textbf{54.9} \\
			\hline
			
			Deeplabv3+ \cite{chen2018encoder} & \multirow{7}{*}{ShuffleNetV2}&33.1 &25.6 &22.2 &28.6 &23.3 && 31.3 &22.2 &25.9 &28.5 &&38.1 &43.3 &36.5 &25.3 && 38.8 &25.0 &47.3 &46.8 && 40.9 &25.1 &46.6 &39.9 \\
			
			IBN \cite{pan2018two} & &35.5 &27.1 &31.8 &34.9 &25.6 && 32.7 &22.8 &26.7 &30.4 &&41.9 &46.9 &\underline{40.9} &26.5 &&40.6 &25.9 &48.6 &48.6 && 42.3 &26.6 &47.8 &42.1 \\
			
			ISW \cite{choi2021robustnet} & &35.6 &31.0 &32.1 &35.3 &24.3 && \underline{33.7} &22.3 &26.3 &28.8 &&41.9 &47.1 &40.2 &27.1 && 40.7 &26.3 &48.4 &48.7 && 41.5 &25.5 &48.7 &42.4 \\
			
			SAN $\dagger$ \cite{peng2022semantic} & &36.3 &\underline{31.9} &30.2 &34.8 &26.2&& 32.1&22.3&26.2&28.6 &&42.3 &49.7 &38.8 &\underline{27.6} &&40.8 &\underline{28.1} &49.8 &50.0 && \underline{42.8} &28.1 &\underline{51.7} &\underline{43.9} \\
			
			PinMem $\dagger$ \cite{kim2022pin} & &36.1 &29.5 &31.3 &35.4 &\underline{29.1} && 32.5 &23.0 &27.3 &30.5 &&39.9 &48.1 &37.4 &25.9 && 39.8 &\textbf{28.5} &49.0 &48.3 && 41.5 &\underline{30.6} &51.5 &43.7 \\
			
			SHADE $\dagger$ \cite{zhao2022style} & &\underline{37.2} &\textbf{35.4} &\underline{32.3} &\textbf{36.9} &28.7 && \textbf{35.4}&\underline{23.4} &\textbf{28.4}&\textbf{34.0} &&\textbf{44.3} &\textbf{51.5} &\underline{40.9} &26.4 && \textbf{43.0} &26.2 &\underline{50.6} &\textbf{51.0} && 41.1 &26.5 &47.2 &40.5 \\
			
			Ours & & \textbf{38.9}& \textbf{35.4}& \textbf{35.9}& \underline{36.3}& \textbf{30.9}&& \textbf{35.4}& \textbf{24.7} & \underline{28.0}& \underline{32.0}&& \underline{43.6}&  \underline{50.8}& \textbf{41.1}& \textbf{28.1}&&   \underline{41.5}&  \textbf{28.5}&  \textbf{51.6} &  \underline{50.5}&&  \textbf{46.1} &  \textbf{31.4}&  \textbf{56.9}& \textbf{48.8} \\
			\hline
			
			Deeplabv3+ \cite{chen2018encoder} & \multirow{7}{*}{MobileNetV2}&33.5 &25.9 &25.7 &26.5 &24.0 && 30.1&20.3&22.8&27.5&&40.2 &44.2 &37.8 &25.5 && 40.1 &26.6 &48.2 &48.7 && 39.1 &27.9 &47.5 &41.6 \\
			
			IBN \cite{pan2018two} & &35.9 &30.1 &27.7 &27.1 &25.0 && 34.3&20.7&23.6&29.9&&45.0 &46.9 &41.1 &\underline{27.6} && 42.7 &27.3 &\underline{52.9} &51.3 && 43.0 &29.2 &50.7 &42.4 \\
			
			ISW \cite{choi2021robustnet} & &36.4 &30.9 &30.1 &30.7 &24.4 && 34.0&23.5&26.2&29.6&&45.2 &49.7 &\underline{41.2} &27.2 && 42.8&28.0 &51.5 &51.6 &&39.7 &29.3 &51.1 &41.8 \\
			
			SAN $\dagger$ \cite{peng2022semantic} & &37.2 &32.5 &27.6 &30.8 &\underline{30.4}&& 32.8&21.9&26.6&32.9 &&45.8 &50.1 &\underline{41.2} &26.7 && 42.5 &27.3 &51.9 &50.8 && \underline{43.8} &\underline{31.1} &52.5 &45.1 \\
			
			PinMem $\dagger$ \cite{kim2022pin} & &37.6 &32.2 &29.0 &31.5 &26.5 && 34.9 &23.7 &27.9 &32.1 &&\underline{46.2} &51.9 &40.4 &26.8 && 43.5 &\textbf{31.2} &49.8 &51.5 && 43.6 &\underline{31.1} &\underline{52.8} &44.5 \\
			
			SHADE $\dagger$ \cite{zhao2022style} & &\underline{38.3} &\underline{34.4} &\underline{32.4} &\underline{32.4} &27.1 && \textbf{36.2}&\underline{23.8} &\underline{28.2} &\underline{33.5} &&46.1 &\underline{53.2} &\underline{41.2} &27.2 && \underline{43.8} &\underline{30.8} &52.6 &\underline{52.0} && 42.3 &31.0 &52.0 &\underline{45.9} \\
			
			Ours & & \textbf{40.8}&  \textbf{36.9}& \textbf{33.7}& \textbf{36.7}& \textbf{31.5}&& \underline{35.5}& \textbf{23.9}& \textbf{28.4}& \textbf{34.3}&& \textbf{48.4} & \textbf{54.2} & \textbf{43.2} & \textbf{27.3} && \textbf{44.6}  & 30.5 & \textbf{53.7} & \textbf{53.3} && \textbf{50.9} & \textbf{33.3} & \textbf{62.7} & \textbf{53.1} \\
			\hline
			
		\end{tabular}
	}
\end{table*}

Specifically, first, the similarity matrix of the prototypes between the source and augmented domains is obtained by using cosine similarity, which is denoted as follows:
\begin{equation}
S = \frac{Z_x^p \cdot Z_{A(x)}^p}{||Z_x^p||\cdot||Z_{A(x)}^p||}
\label{cosine_sim}
\end{equation}
where the similarity matrix $S \in \mathcal{R}^{C\times C}$. In a perfect case, the similarity of the prototypes of the same class between the two domains is the bigger the better and the similarity of the prototypes of different classes between the two domains is the smaller the better. The hard-aligned prototypes are in the opposite case. To calibrate the weights of these prototypes, the hard-weighted matrix $H$ is generated as follows:
\begin{equation}
H = abs(M - E - S)
\label{hard_weight}
\end{equation}
where $M$ is an all-ones matrix and $E$ is an identity matrix. The $abs()$ is to compute the absolute value of a number. In the hard-weighted matrix, a small value means the prototypes of the corresponding class are hard-aligned. Finally, the hard-weighted matrix is utilized to calibrate the prototypes of the augmented domains $Z_{A(x)}^{p}$ during the PCL. Thus, the HPCL is denoted as follows by rewriting Eq. (\ref{contrast_sim_compute}):
\begin{equation}
\mathcal{L}_{hpcl} = -\sum_{i=1}^{C}y_s^i\log \frac{exp(((Z_{A(x)}^{p_i} \times H_{i,i})\cdot Z_x^{c_i}) /\tau_h)}{\sum_{i\neq k}^{C} exp(((Z_{A(x)}^{p_k} / H_{i,k})\cdot Z_x^{c_i}) /\tau_h)}
\label{loss:hpcl}
\end{equation}
where $\tau_h$ is the temperature parameter. When the prototypes are hard-aligned, the $Z_{A(x)}^{p}$ is weighted with a small value. Thus, the numerator and denominator of the $\mathcal{L}_{hpcl}$ are respectively decreased and increased, which means the loss $\mathcal{L}_{hpcl}$ is increased to optimize the class-wise features for aligning these hard-aligned prototypes.

\subsection{The training loss}
The overall objective of our approach can be formulated as follows:
\begin{equation}
\mathcal{L}= \mathcal{L}_{seg} + \lambda_1 \mathcal{L}_{upcl} + \lambda_2 \mathcal{L}_{hpcl}
\label{all:loss}
\end{equation}
where $\lambda_i$ is weights coefficients. In particular, during the training process, we freeze the weights of the segmentation network to extract the augmented features $Z_{A(x)}$. 

\subsection{Behavior in Different Situations} 
We further discuss the behavior of the proposed approach in different situations. Specifically, we divide the prototype discrepancy in different domains into the following three situations: i) the prototypes of different domains are similar or even the same; ii) the prototypes of different domains may have discrepancies; iii) the prototypes are far away in different domains. In the first situation, all prototypes are assigned with big weights when the prototypes of the source and augmented domains are similar. In particular, the weights approximately equal 1 when the prototypes of different domains are the same. In this situation, the proposed approach can be viewed as the conventional prototypical contrastive learning (PCL). In the second situation, an uncertainty probability matrix and a hard-weighted matrix are generated to calibrate the weights of the prototypes which have a big gap in the source and augmented domains during the prototypical contrastive learning. In the last situation, there is a big gap between the prototypes in the source and augmented domains. All prototypes are assigned with small weights, which means that the class-wise features are pushed away from the prototypes of the source domain to some extent. We argue that it can prevent the model from overfitting the prototypes of the source domain.

\begin{figure*}[!t]
	\centering
	\subfigure[Input image]{
		\begin{minipage}{0.145\textwidth}
			\includegraphics[width=2.6cm,height=1.2cm]{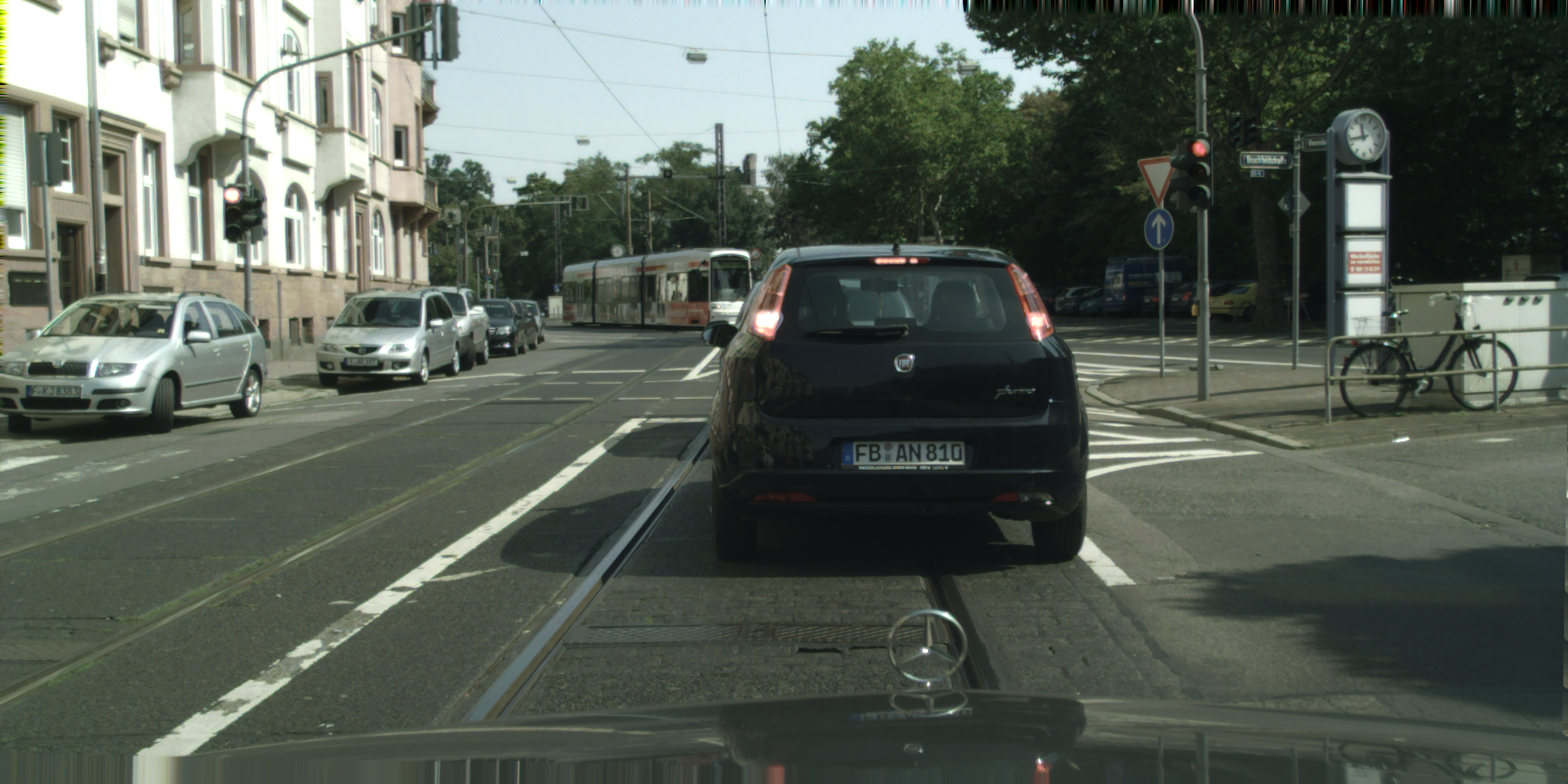} 
			\includegraphics[width=2.6cm,height=1.2cm]{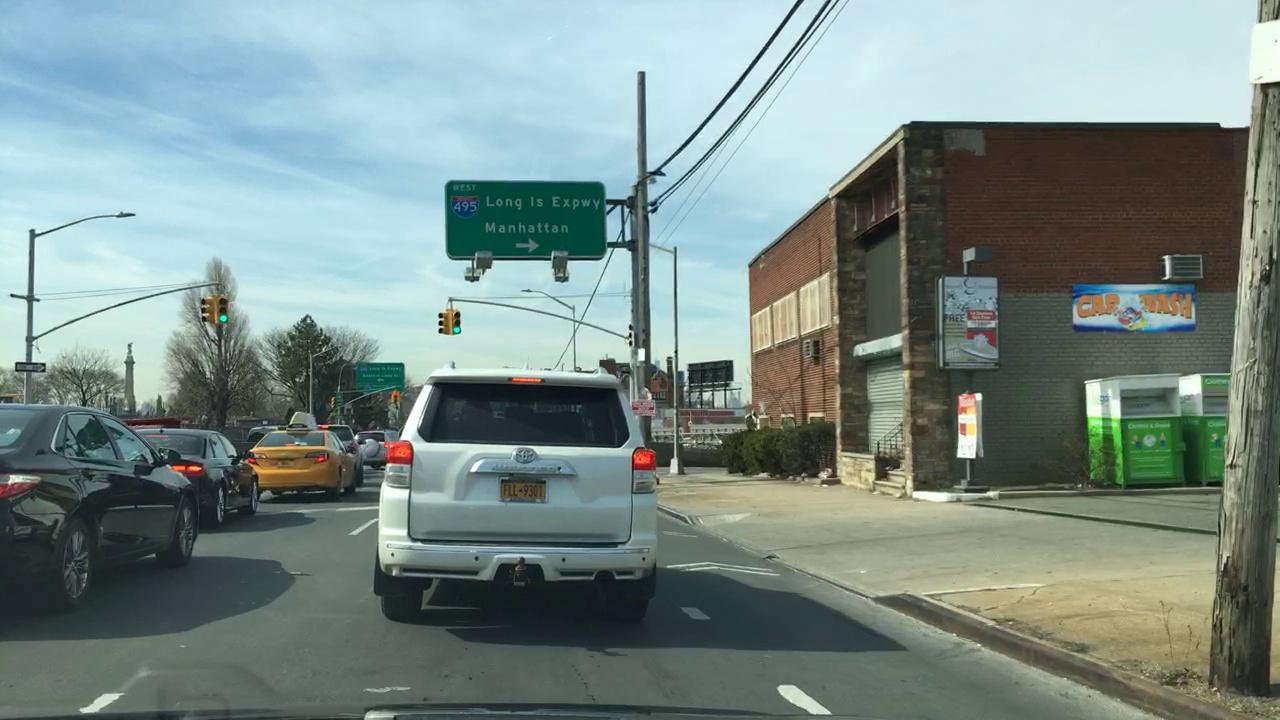}
			\includegraphics[width=2.6cm,height=1.2cm]{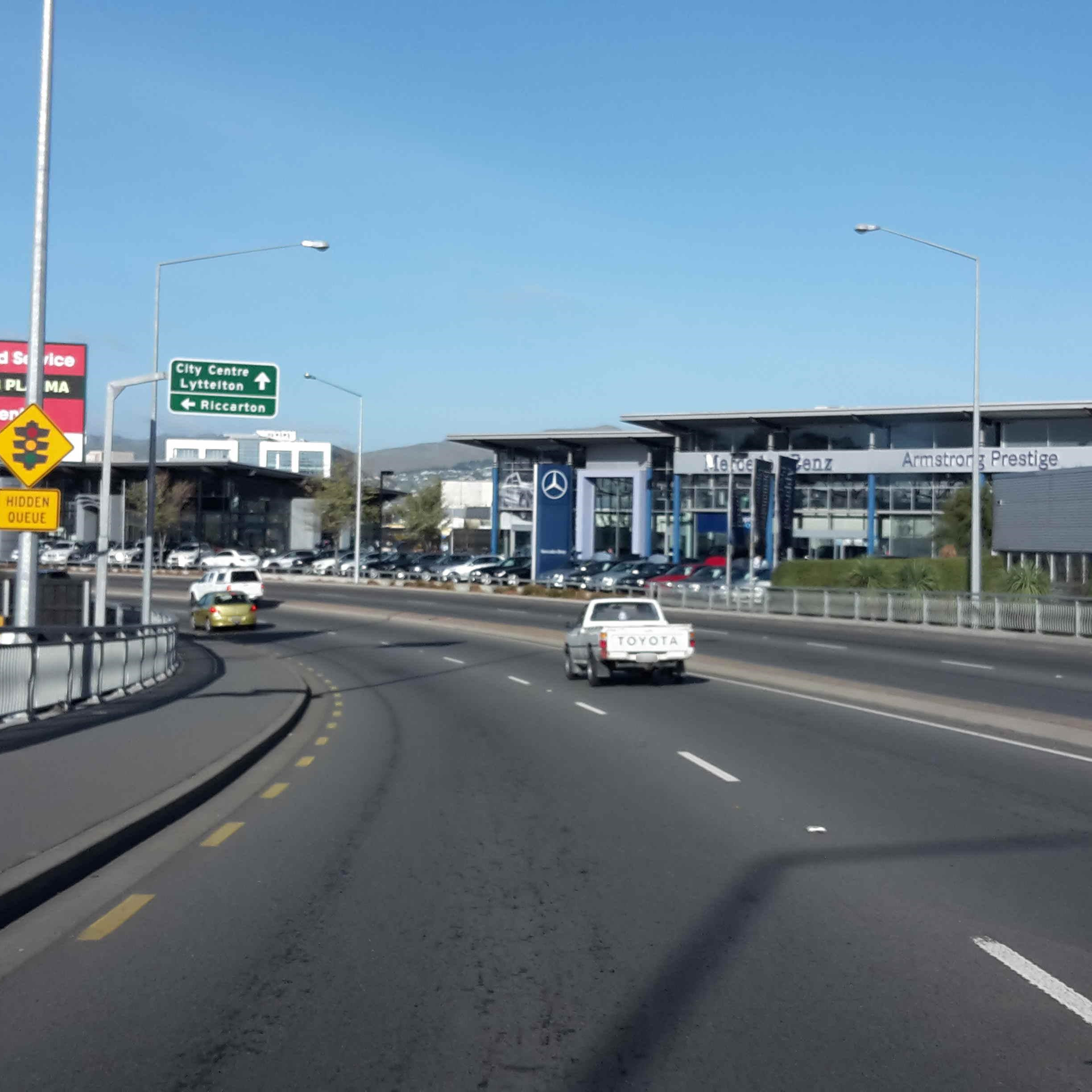} 
			\includegraphics[width=2.6cm,height=1.2cm]{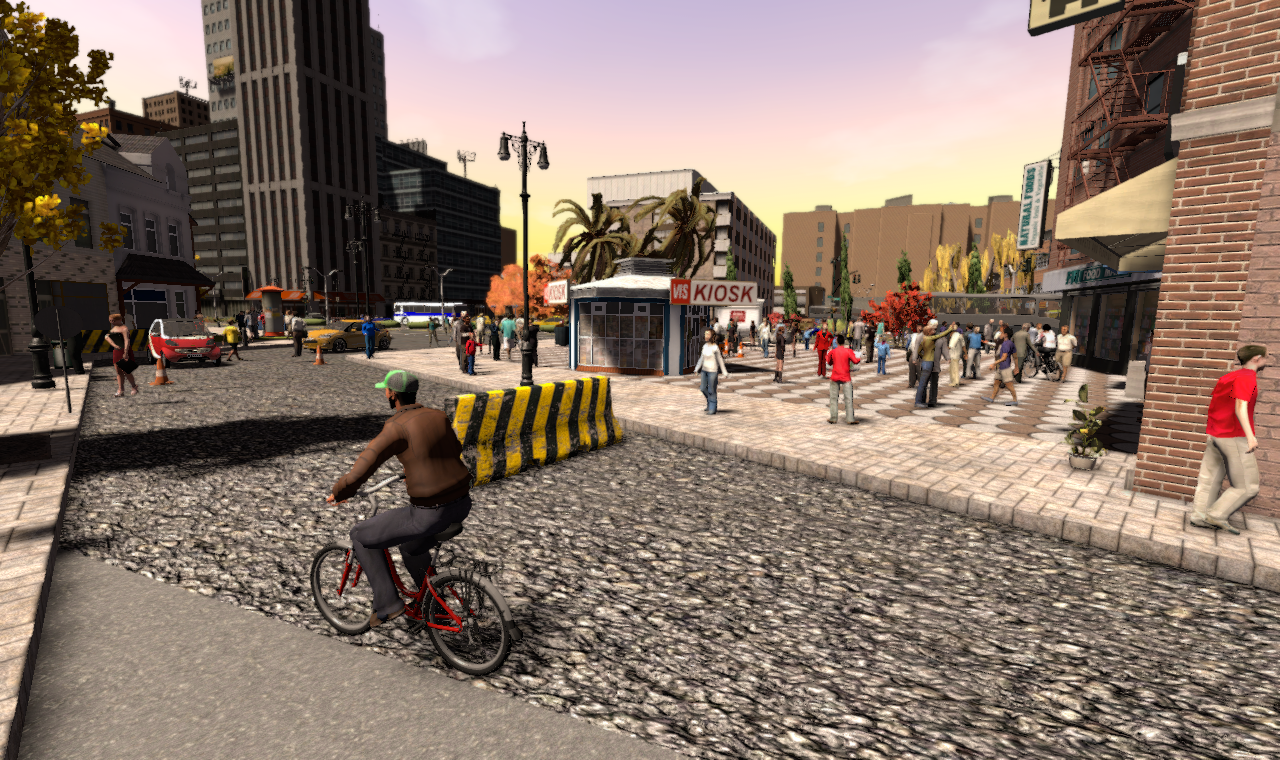} 
		\end{minipage}
	}
	\subfigure[Ground truth]{
		\begin{minipage}{0.145\textwidth}
			\includegraphics[width=2.6cm,height=1.2cm]{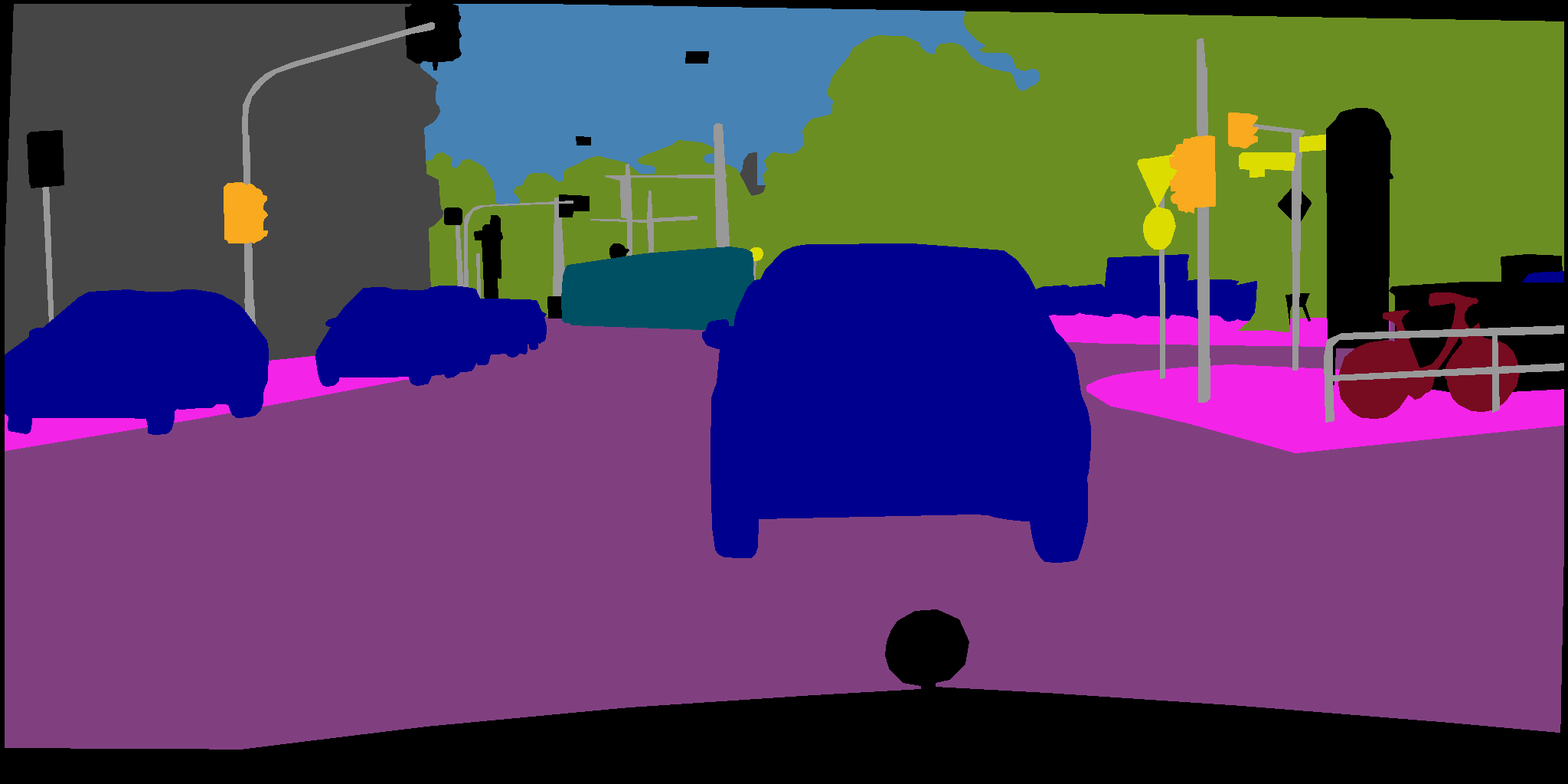}
			\includegraphics[width=2.6cm,height=1.2cm]{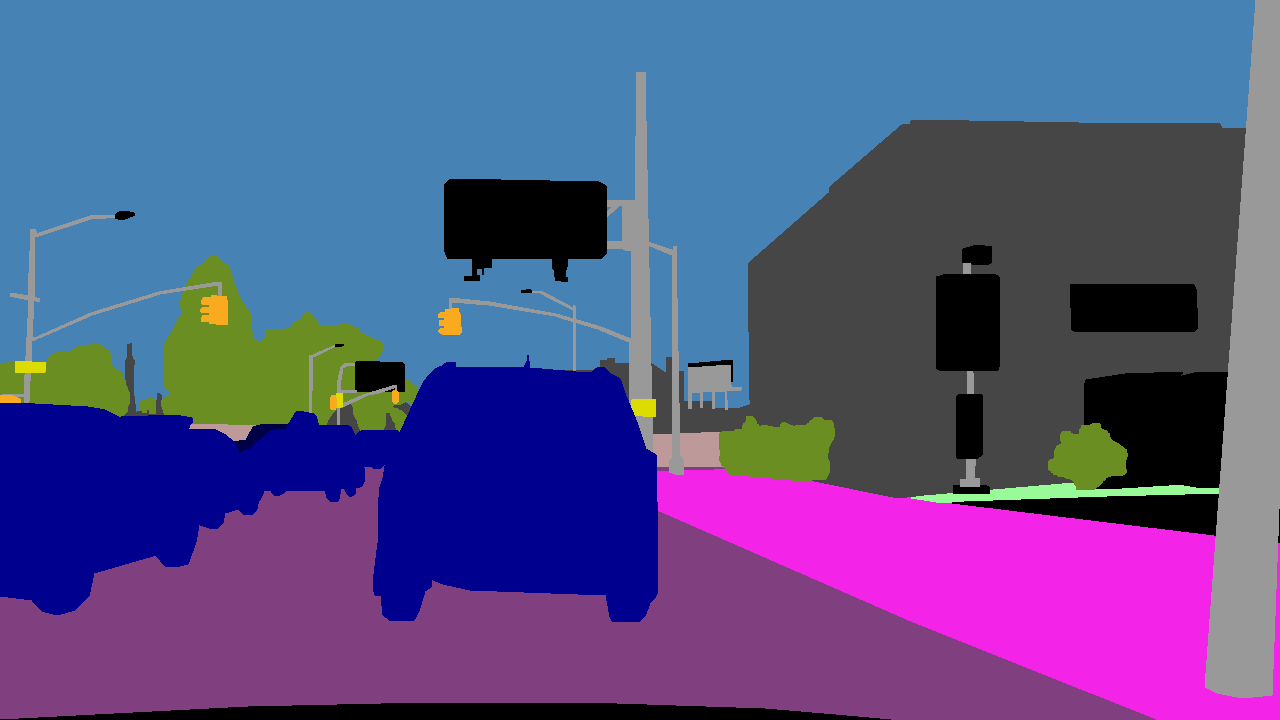} 
			\includegraphics[width=2.6cm,height=1.2cm]{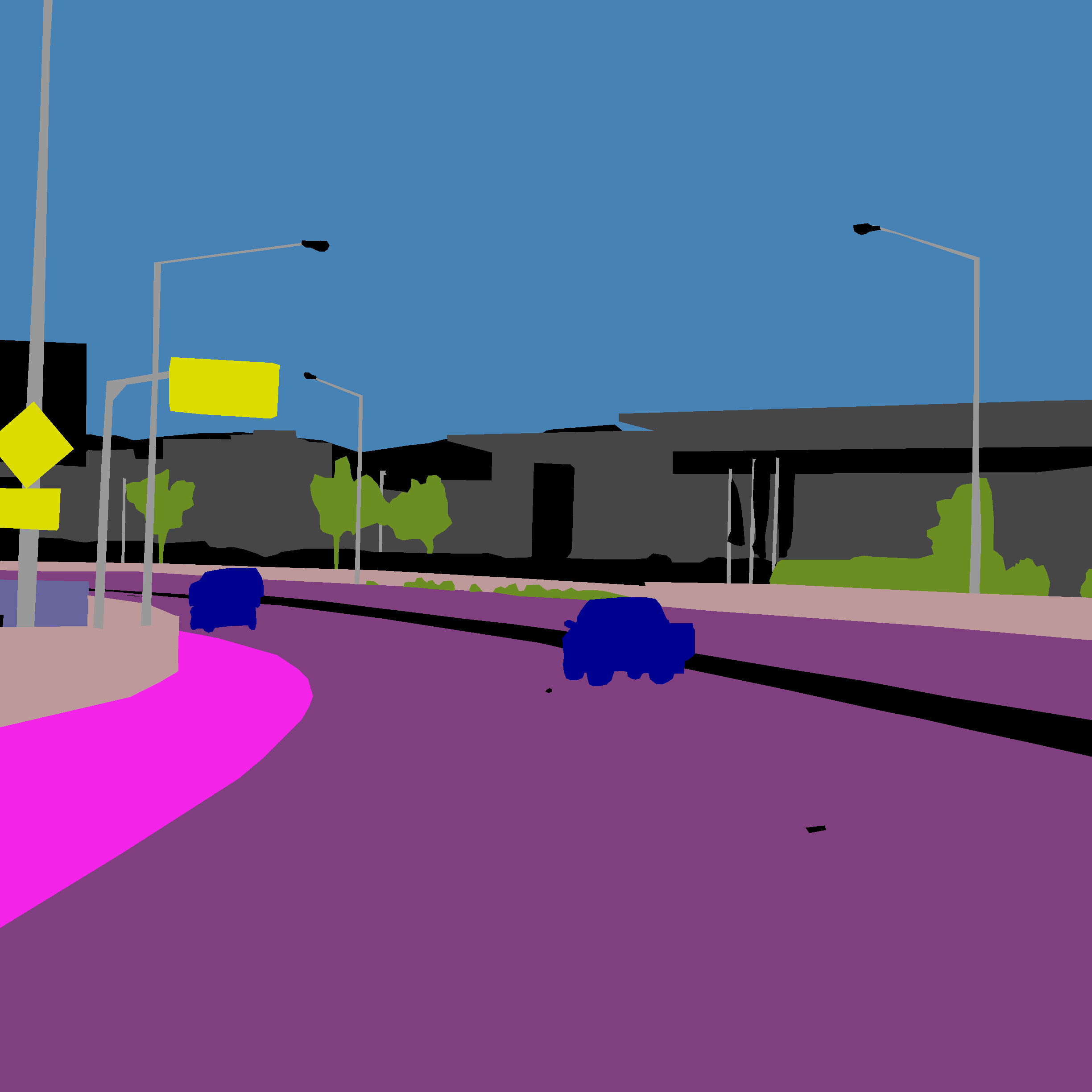}
			\includegraphics[width=2.6cm,height=1.2cm]{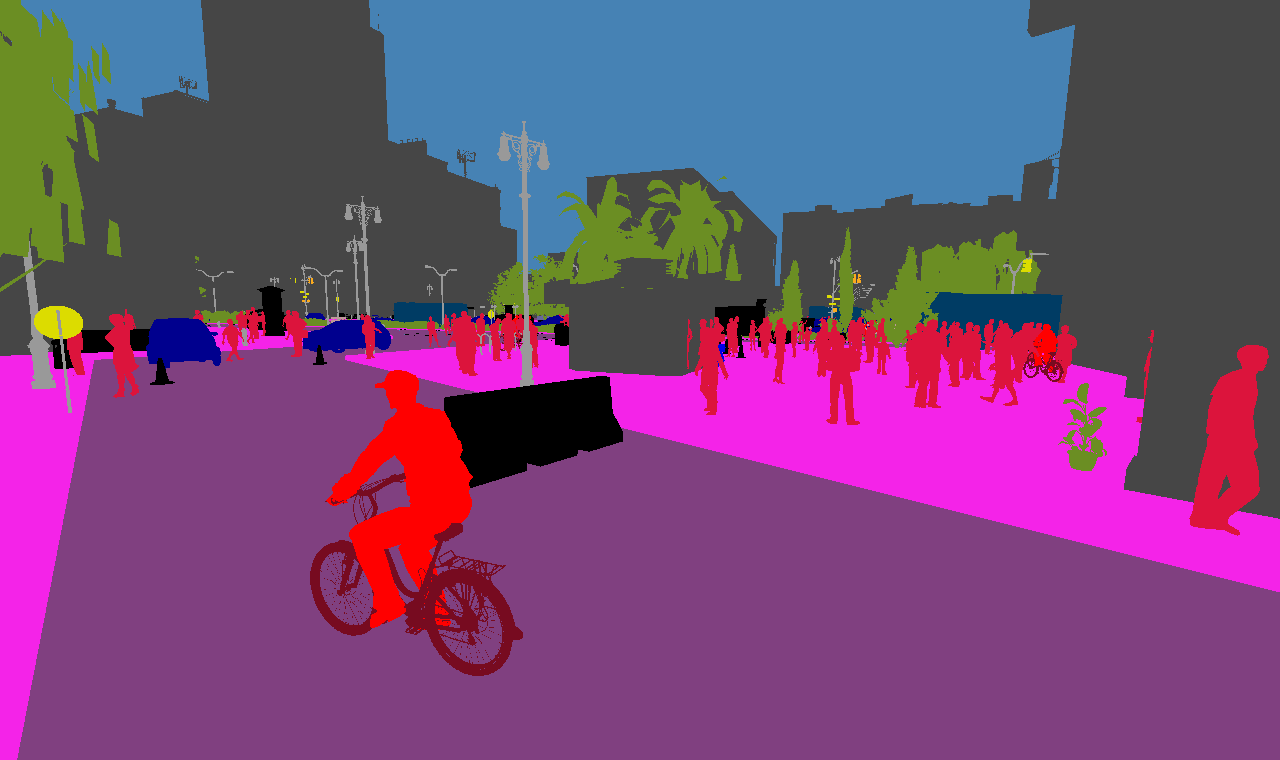} 
		\end{minipage}
	}
	\subfigure[ISW \cite{choi2021robustnet}]{
		\begin{minipage}{0.145\textwidth}
			\includegraphics[width=2.6cm,height=1.2cm]{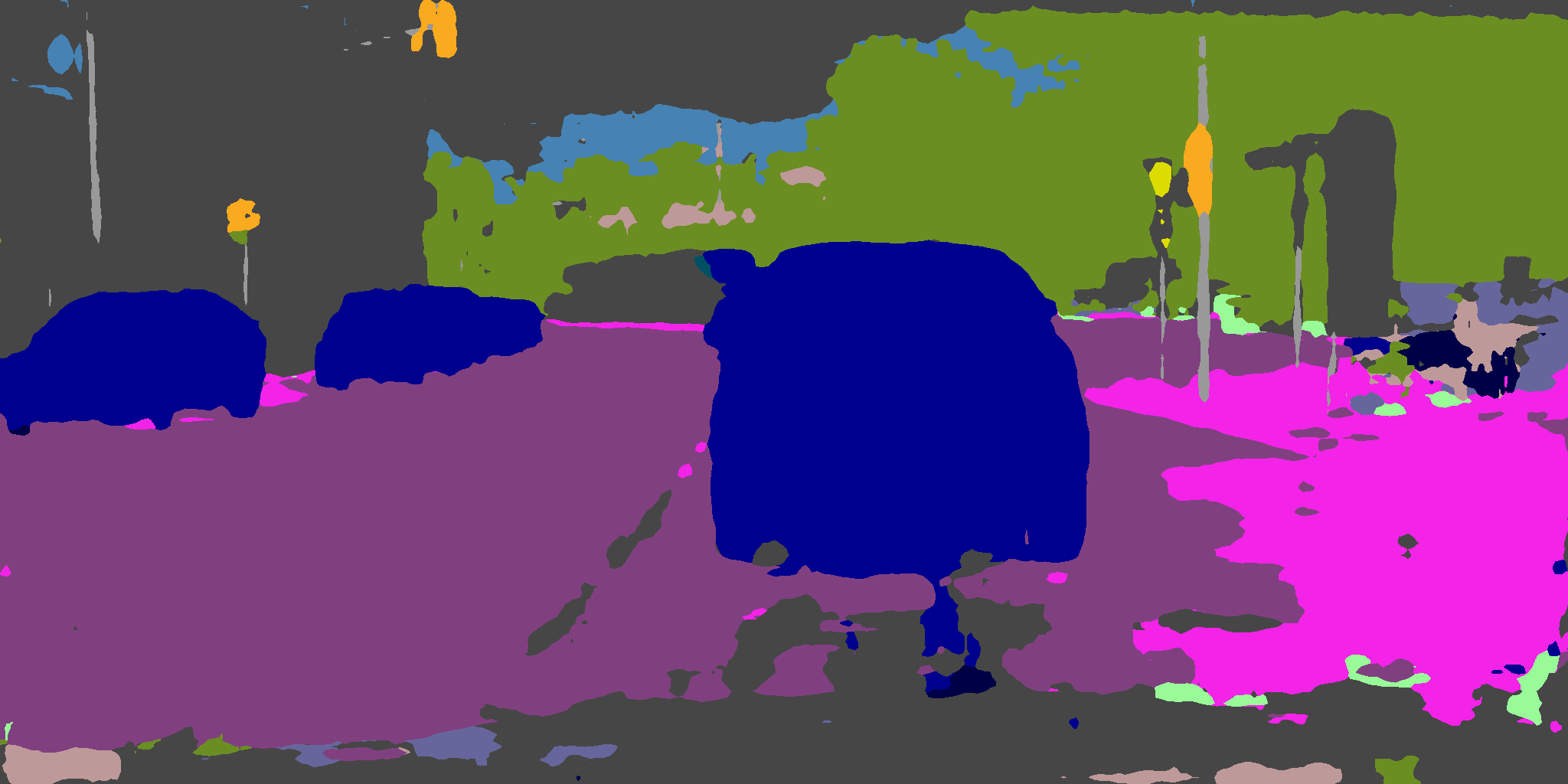}
			\includegraphics[width=2.6cm,height=1.2cm]{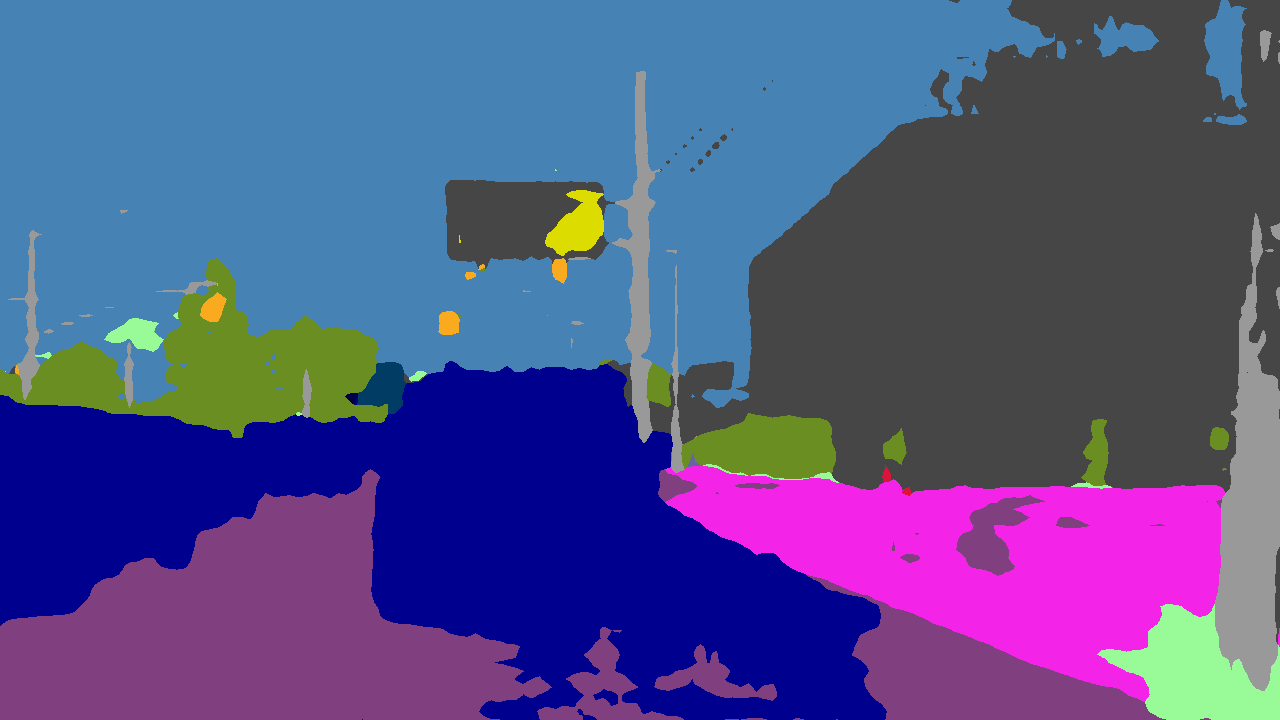} 
			\includegraphics[width=2.6cm,height=1.2cm]{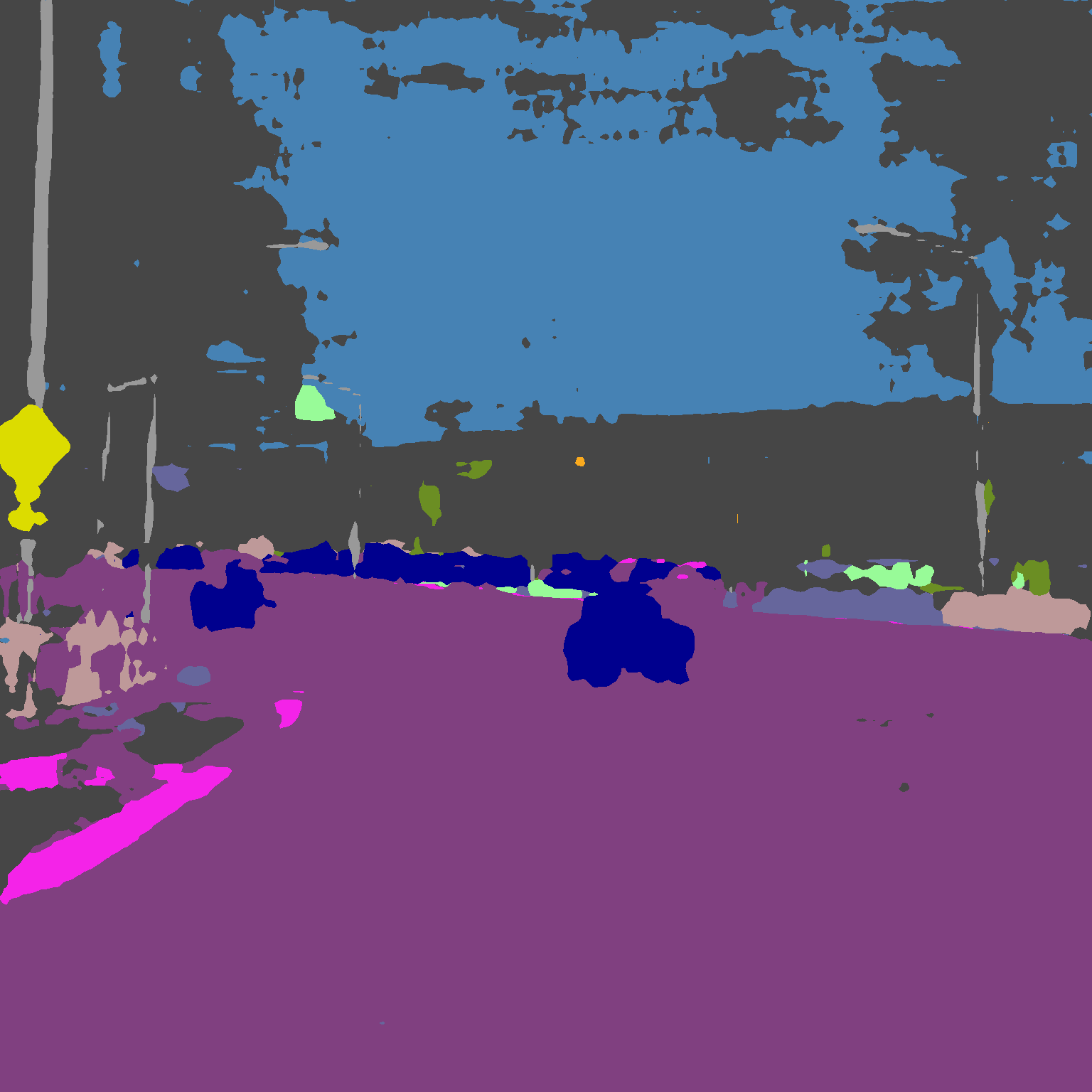}
			\includegraphics[width=2.6cm,height=1.2cm]{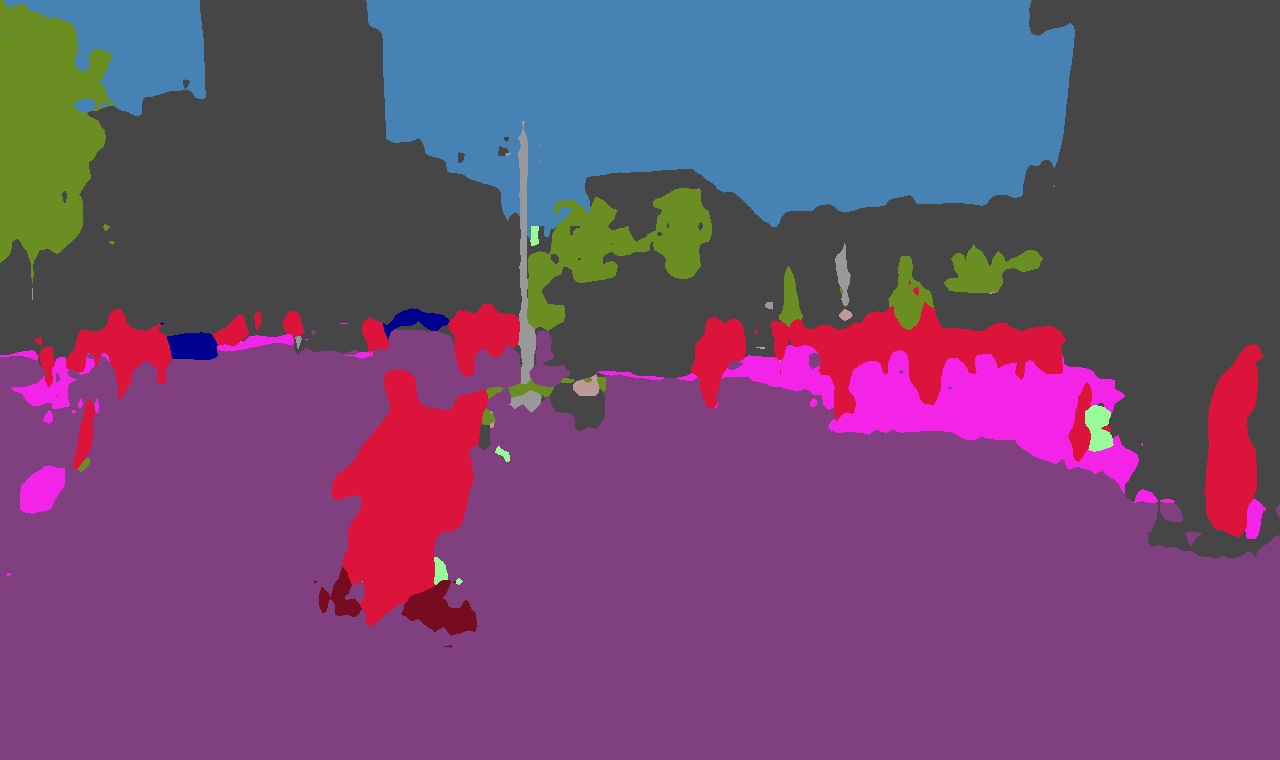}
		\end{minipage}
	}
	\subfigure[PinMem \cite{kim2022pin}]{
		\begin{minipage}{0.145\textwidth}
			\includegraphics[width=2.6cm,height=1.2cm]{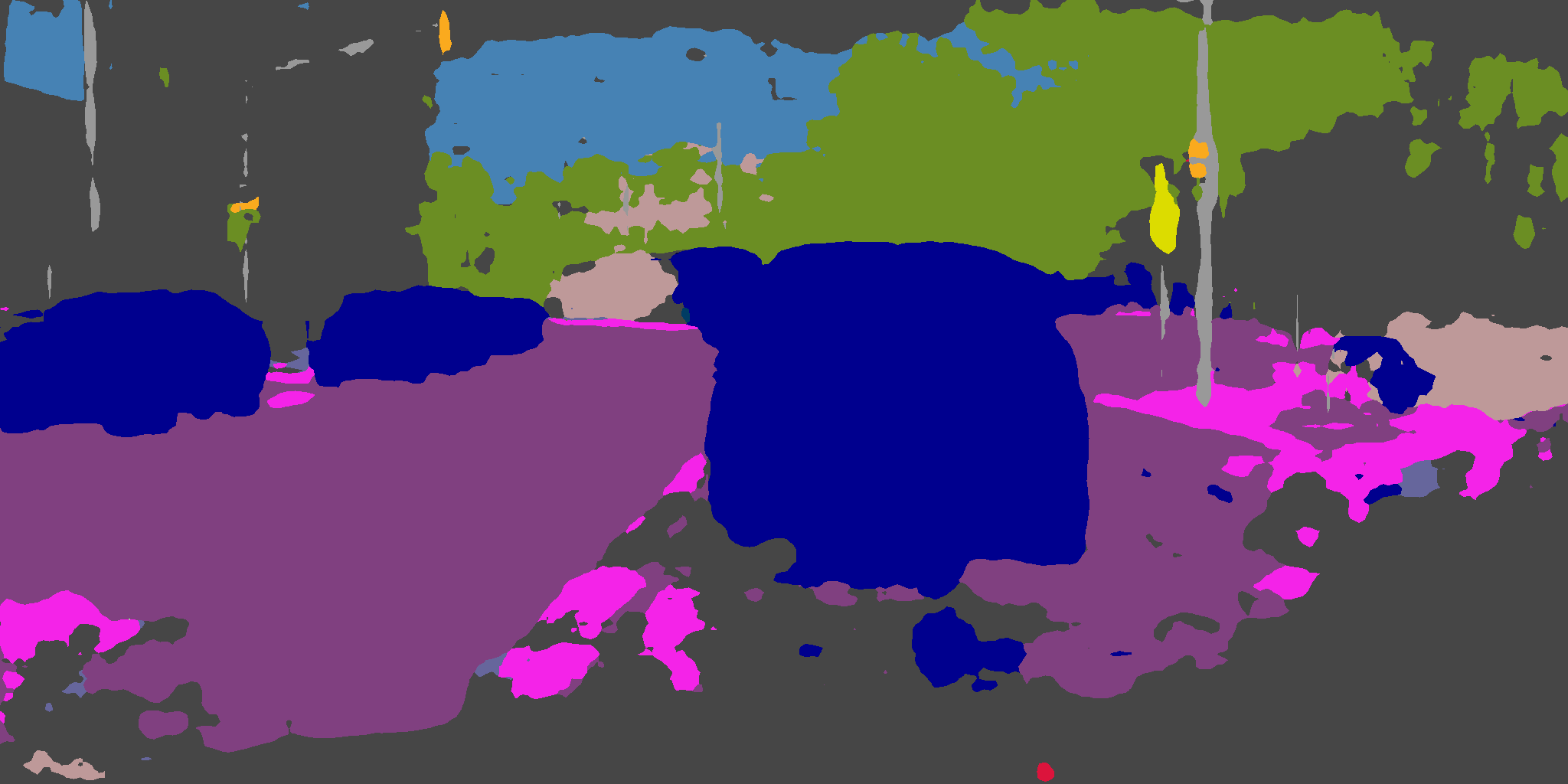}
			\includegraphics[width=2.6cm,height=1.2cm]{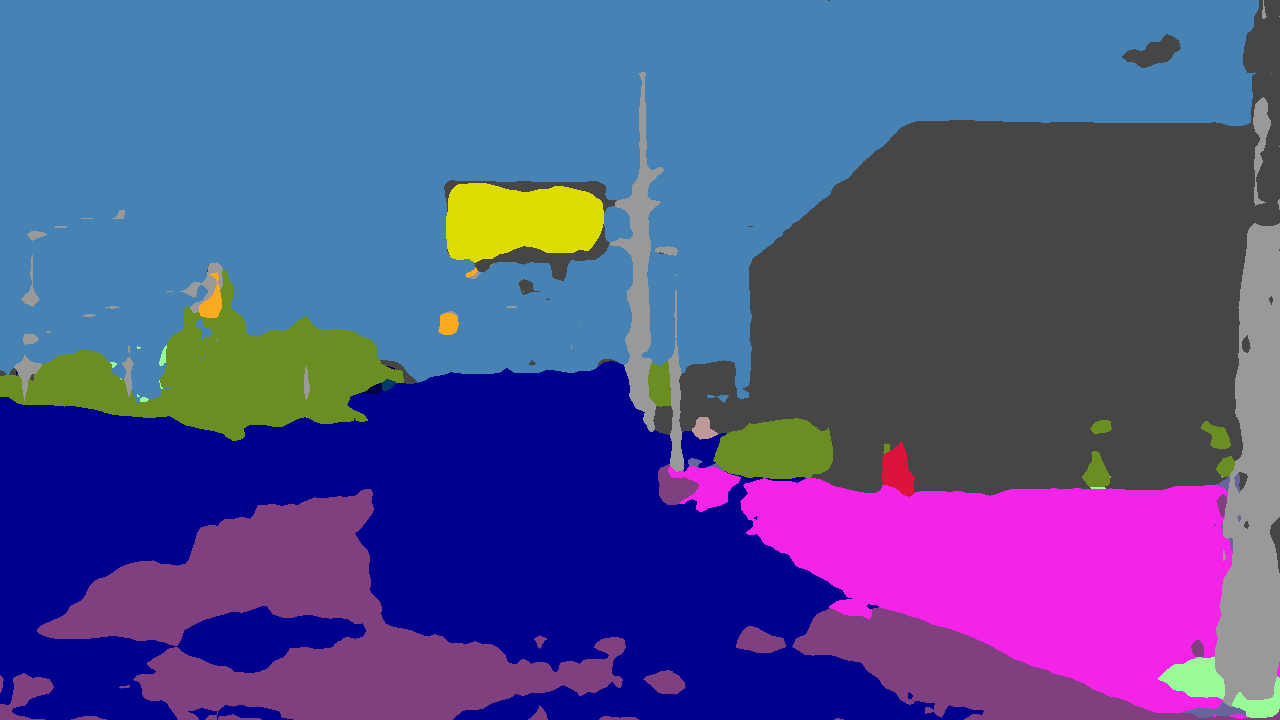} 
			\includegraphics[width=2.6cm,height=1.2cm]{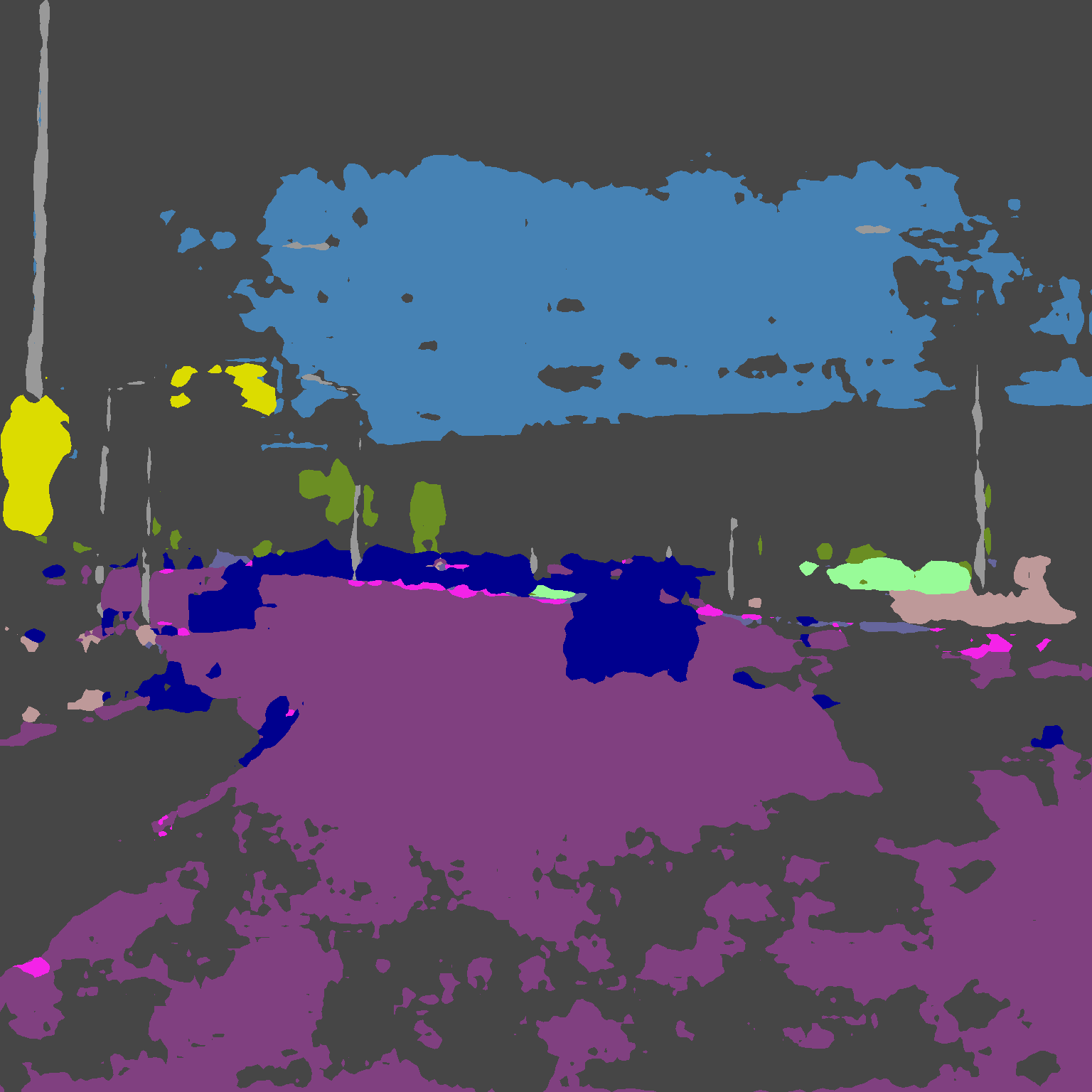}
			\includegraphics[width=2.6cm,height=1.2cm]{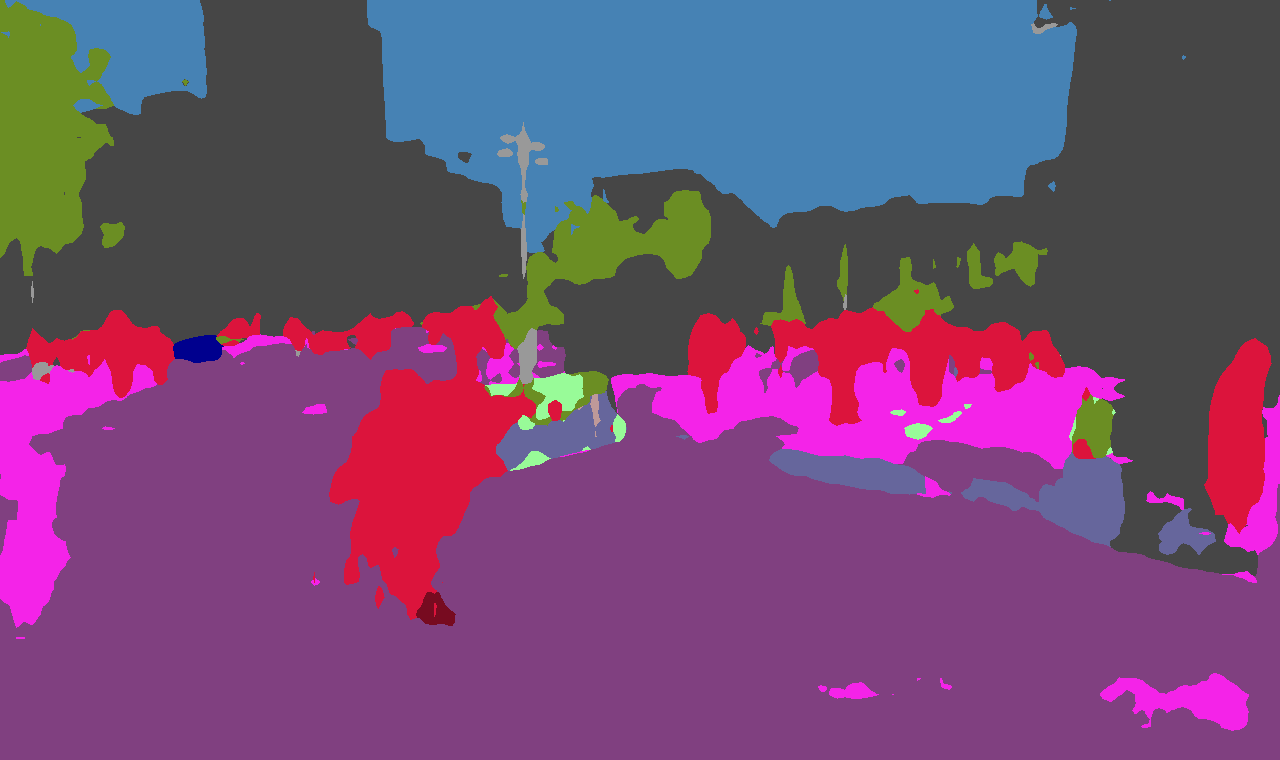}
		\end{minipage}
	}
	\subfigure[SHADE \cite{zhao2022style}]{
		\begin{minipage}{0.145\textwidth}
			\includegraphics[width=2.6cm,height=1.2cm]{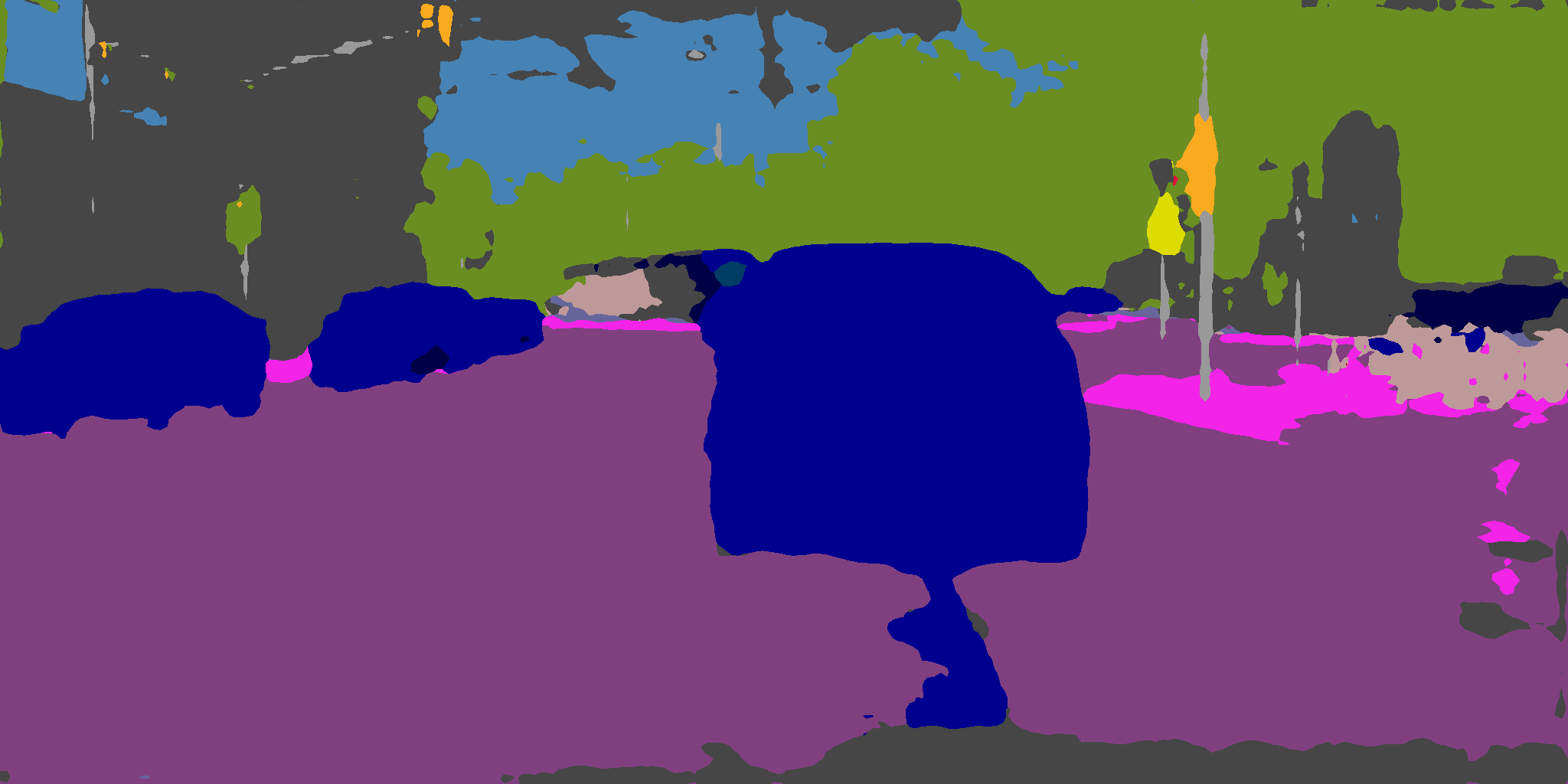}
			\includegraphics[width=2.6cm,height=1.2cm]{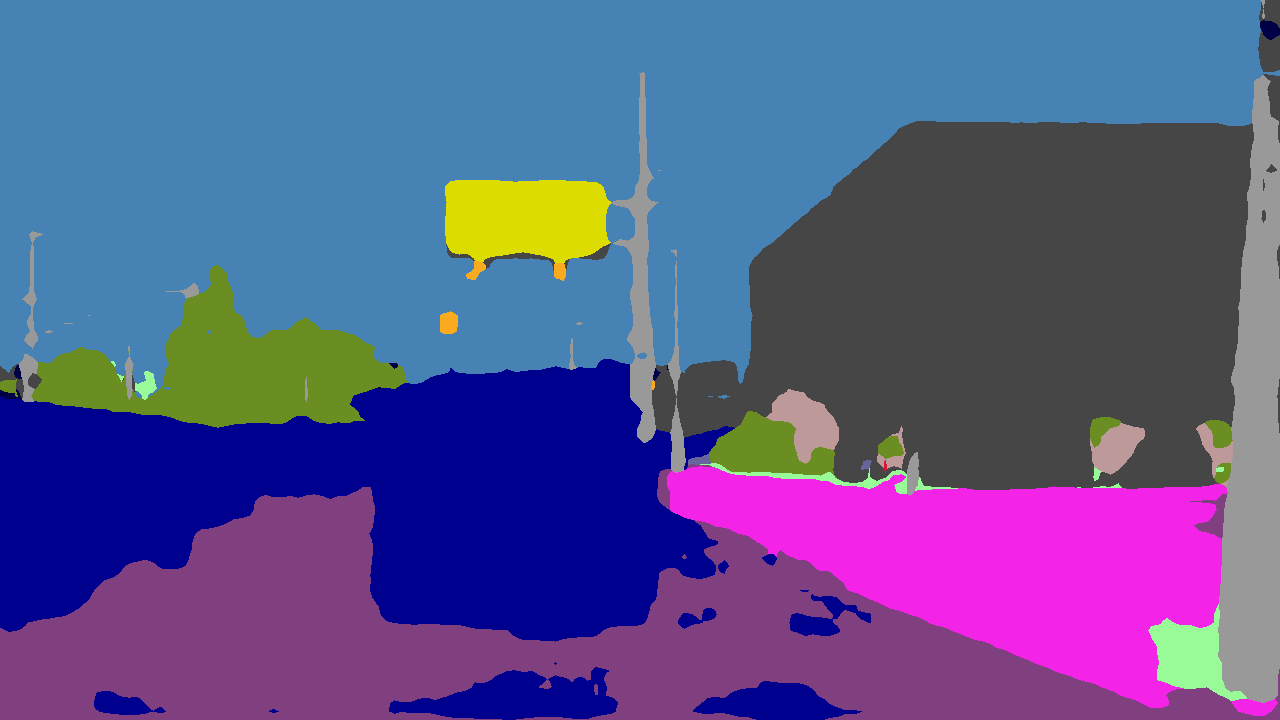} 
			\includegraphics[width=2.6cm,height=1.2cm]{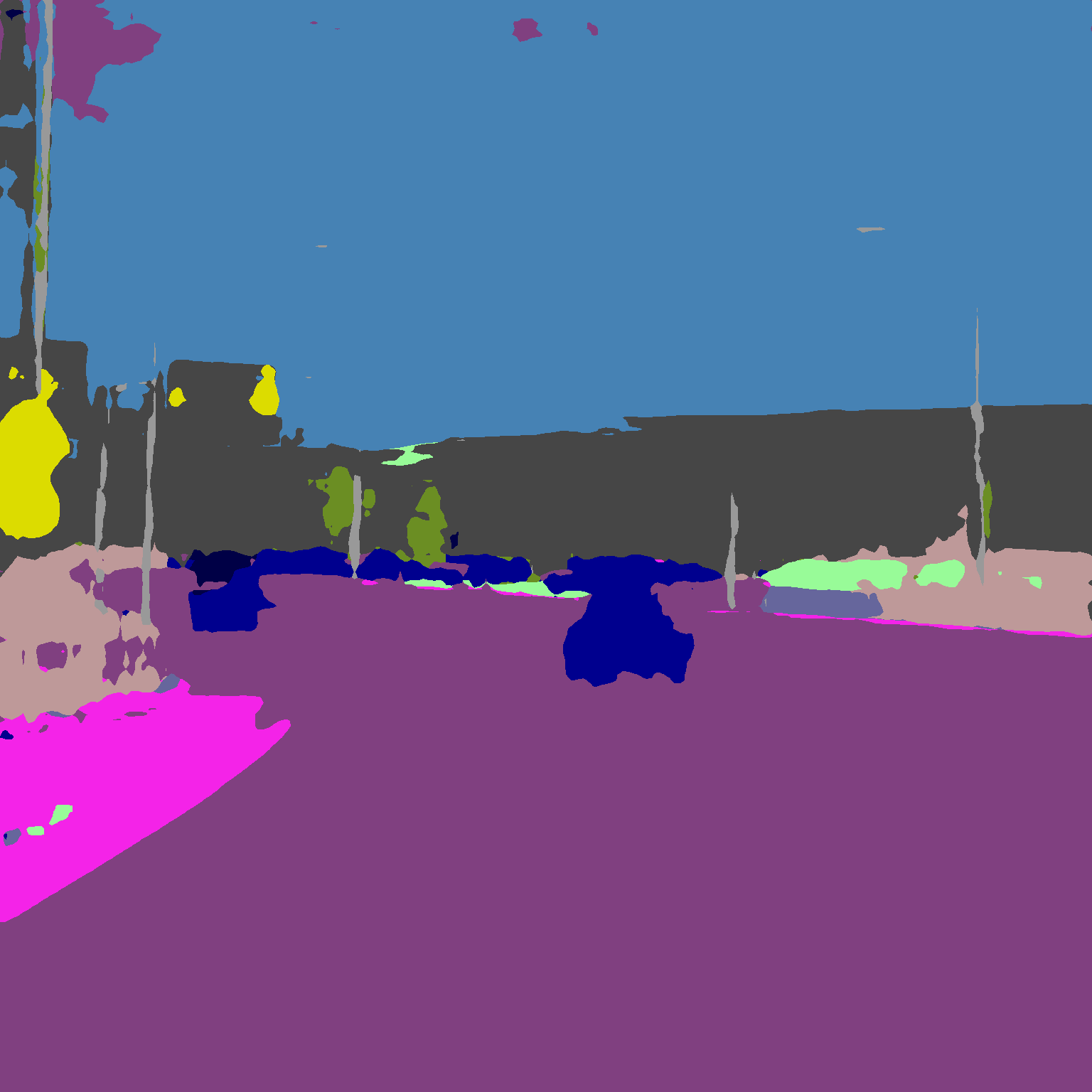}
			\includegraphics[width=2.6cm,height=1.2cm]{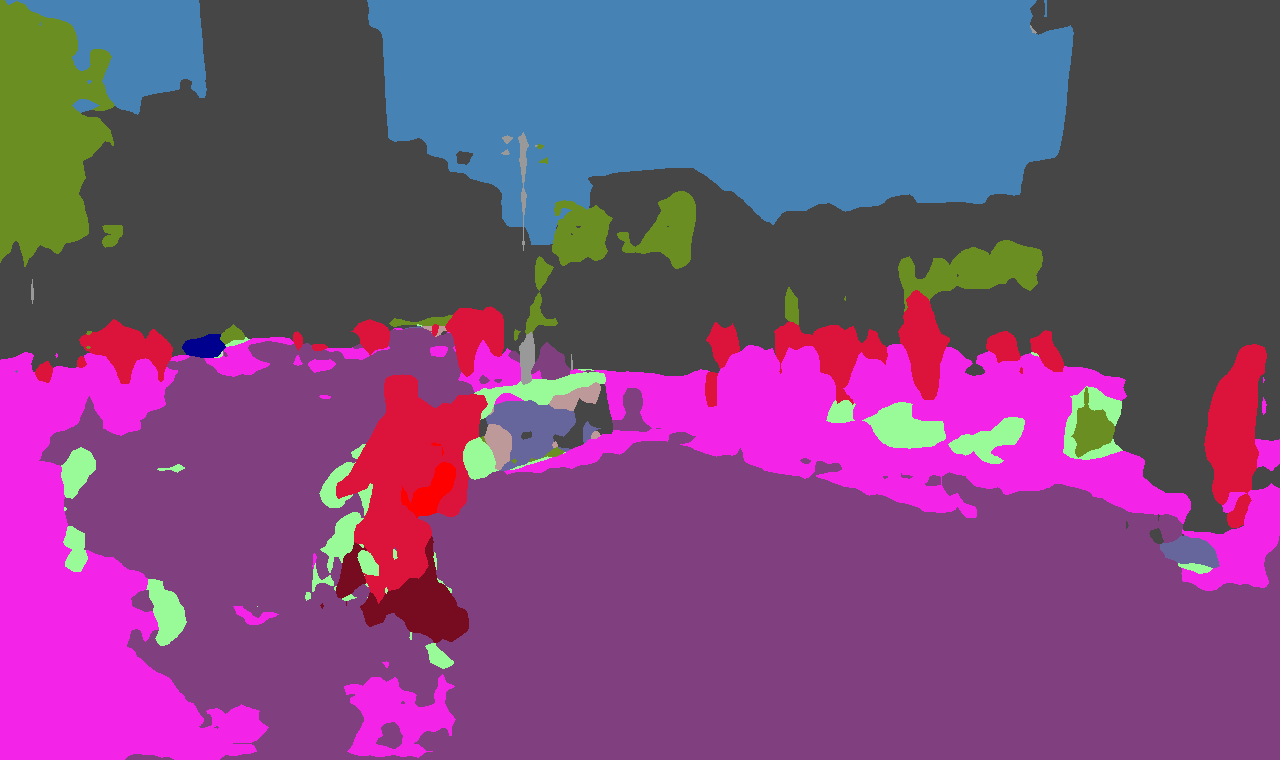}
		\end{minipage}
	}
	\subfigure[Ours]{
		\begin{minipage}{0.145\textwidth}
			\includegraphics[width=2.6cm,height=1.2cm]{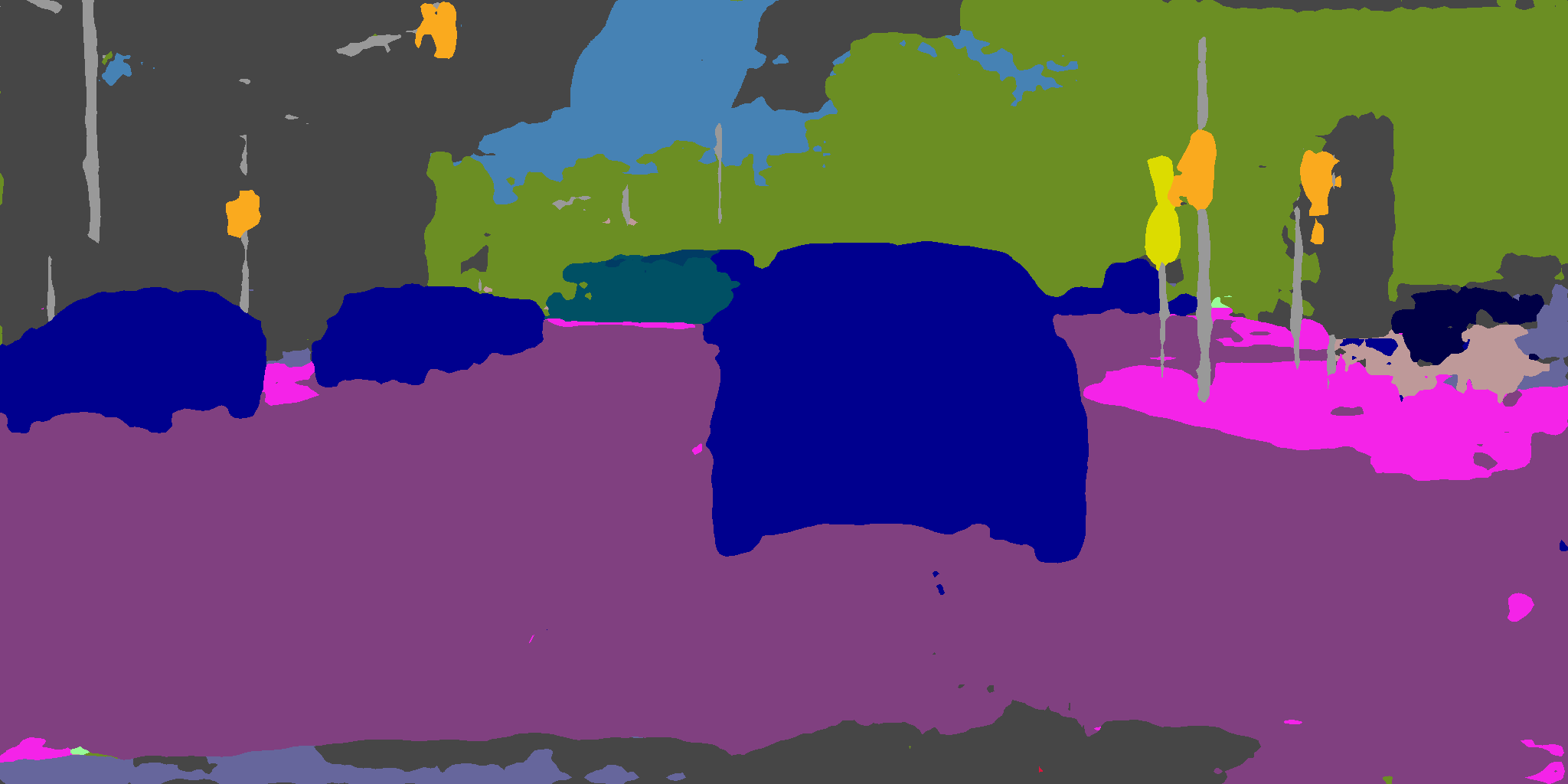}
			\includegraphics[width=2.6cm,height=1.2cm]{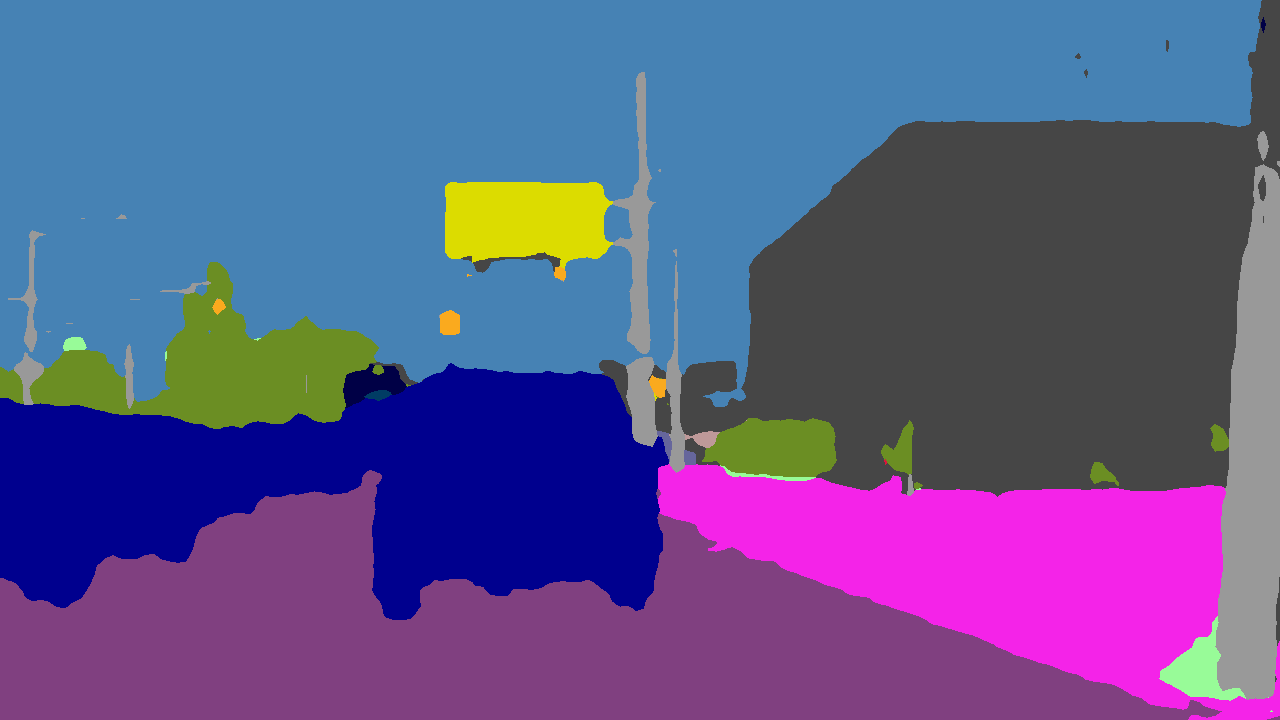}
			\includegraphics[width=2.6cm,height=1.2cm]{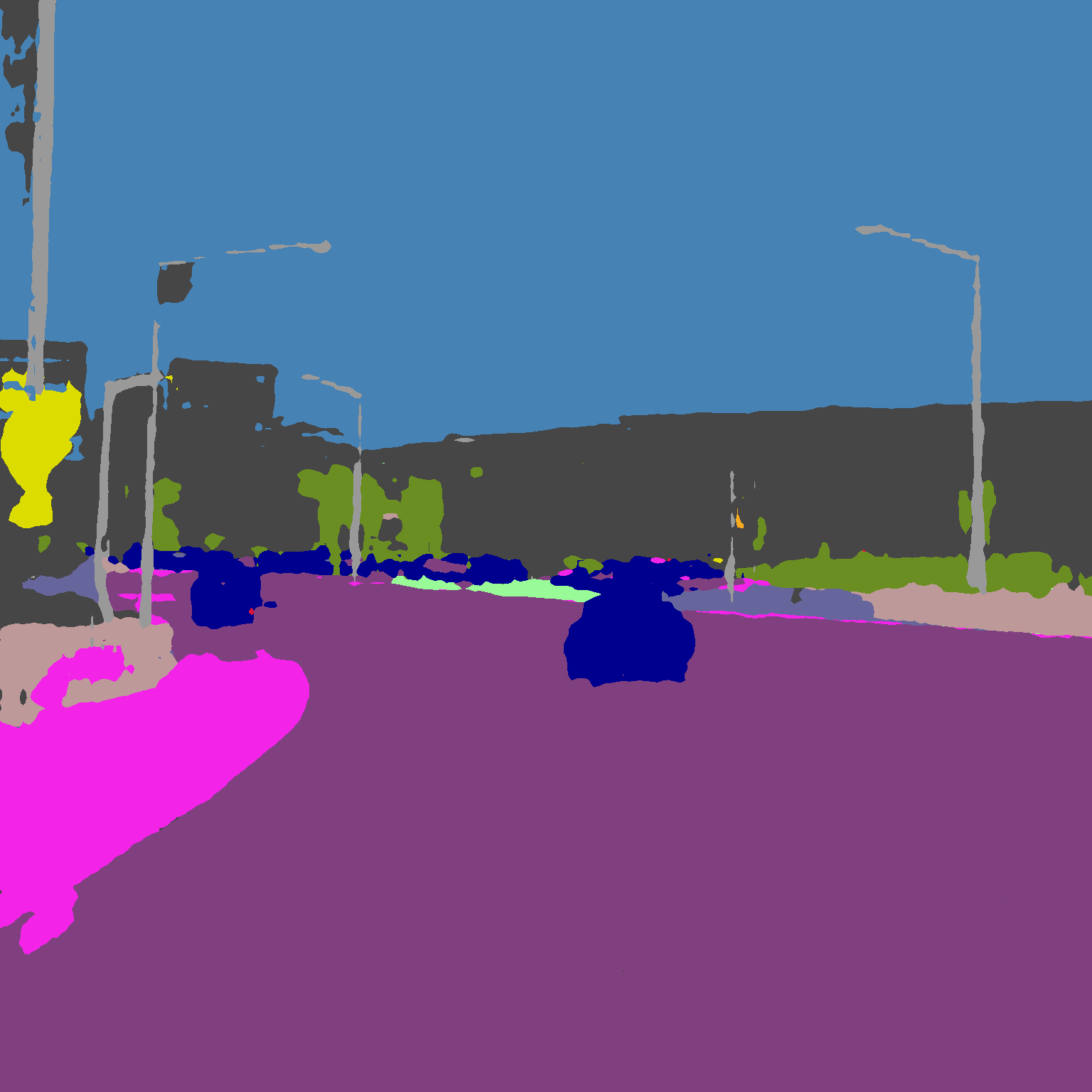}
			\includegraphics[width=2.6cm,height=1.2cm]{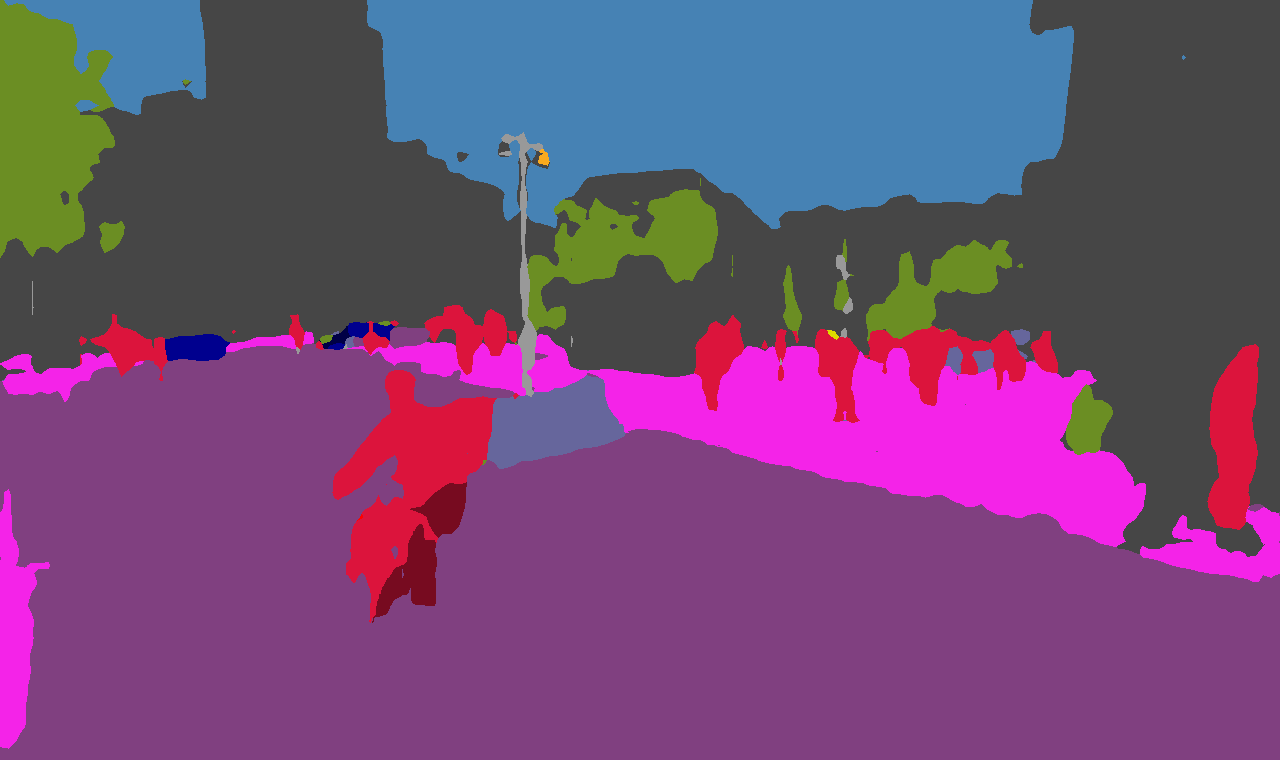}
		\end{minipage}
	}
	\caption{Visualization comparison between recent methods and our proposed approach in the G$\rightarrow$\{C, B, M, S\} task. The visualization results from four datasets (including Cityscapes, BDD, Mapillary, and SYNTHIA) are respectively shown in four rows.}
	\label{fig:vis_G}
\end{figure*}

\section{Experiments}
\subsection{Datasets and Implementation Details} 
Five datasets are tested on the proposed approach, including GTA5 (G) \cite{richter2016playing}, SYNTHIA (S) \cite{ros2016synthia}, Cityscapes (C) \cite{cordts2016cityscapes}, BDD (B) \cite{yu2020bdd100k}, and Mapillary (M) \cite{neuhold2017mapillary}. Our approach is trained on single or several datasets and evaluated on other datasets. In experiments, the ISW \cite{choi2021robustnet} is adopted as baseline. The ResNet-50 \cite{he2016deep}, ShuffleNetV2 \cite{ma2018shufflenet}, and MobileNetV2 \cite{sandler2018mobilenetv2} are utilized in DeepLabV3+ \cite{chen2018encoder} as the segmentation network. We follow the data augmentations of ISW \cite{choi2021robustnet}. Moreover, the weight coefficients $m_p$, $m_a$, $m_u$, $\tau_u$, $\tau_h$, $\lambda_1$, and $\lambda_2$ are set as 0.9, 0.9, 0.9, 0.8, 0.8, 0.1, and 0.01 for all experiments. More implementation details and the weight coefficient experiments are introduced in the Appendix \ref{app_exp_id} and Appendix \ref{app_sm}.

\subsection{Comparison with Single-source Domain Generalization Methods}
We compare the performance of our approach with several recent approaches \cite{pan2018two, pan2019switchable, yue2019domain, peng2021global, choi2021robustnet, peng2022semantic, kim2022pin, lee2022wildnet, zhao2022style}. Our approach achieves superior average performance than these methods in five single-source generalization settings on three backbones, which are shown in Table \ref{tab:all}. For the ResNet-50 backbone, compared with the SOTA methods SAN \cite{peng2022semantic}, PinMem \cite{kim2022pin}, SHADE \cite{zhao2022style}, and WildNet \cite{lee2022wildnet}, our approach respectively achieves the significant average improvement of 6.7\%, 4.2\%, 2.7\%, and 2.6\%. For the ShuffleNetV2 backbone, compared with the SOTA methods SHADE \cite{zhao2022style} and PinMem \cite{kim2022pin}, our approach achieves an average improvement of 1.7\% and 2.8\%. For the MobileNetV2 backbone, compared with the SOTA methods SHADE \cite{zhao2022style} and PinMem \cite{kim2022pin}, our approach achieves an average improvement of 2.5\% and 3.2\%. In addition, the visualization comparisons of segmentation maps between our proposed approach and recent methods are shown in Figure \ref{fig:vis_G}. It shows that the classes of ``road'', ``sidewalk'', and ``car'' are segmented more accurately than recent methods.

\begin{table}[htb]
	\centering
	{
		\caption{Performance comparison in terms of mIoU (\%) between domain generalization methods in the architecture of the ResNet-50 \cite{he2016deep}. The best results are marked in bold and the second-best results are underlined.}
		\label{tab:ms_gs}
		\begin{tabular}{lccccc}
			\hline
			\multirow{2}{*}{Methods}& 
			\multirow{2}{*}{Backbone}& 
			\multirow{2}{*}{Mean}& 
			\multicolumn{3}{c}{Train on G and S}
			\\
			\cline{4-6}
			&&&  $\rightarrow$C&$\rightarrow$B&$\rightarrow$M \\
			\hline
			Deeplabv3+ \cite{chen2018encoder} & \multirow{6}{*}{ResNet-50}&30.8 &35.5 &25.1 &31.9 \\
			
			ISW \cite{choi2021robustnet}& &36.8 &37.7 &34.1 &38.5 \\
			
			MLDG \cite{li2018learning}& &35.5 &38.8 &32.0 &35.6 \\
			
			PinMem \cite{kim2022pin} & &41.8 &44.5 &38.1 &42.7 \\
			
			SHADE \cite{zhao2022style} & &\underline{45.1} &\underline{47.4} &\underline{40.3} &\underline{47.6} \\
			
			Ours& &\textbf{46.7} &\textbf{48.1} &\textbf{42.5} &\textbf{49.4} \\
			\hline
		\end{tabular}
	}
\end{table}

\begin{table*}[!t]
	\caption{Ablation experiments in the ``G$\rightarrow$\{C, B, M, S\}'' generalization setting based on the ShuffleNetV2 backbone. The ``PCL'' indicates the prototypical contrastive learning. The ``UPCL'' and ``HPCL'' mean the uncertainty-weighted and the hard-weighted prototypical contrastive learning.}
	\centering
	\begin{tabular}{llcccccccccc}
		\hline
		\multirow{2}{*}{Backbone}& 
		\multirow{2}{*}{Methods}& 
		\multirow{2}{*}{PCL}&
		\multirow{2}{*}{UPCL}&
		\multirow{2}{*}{HPCL}&
		\multicolumn{5}{c}{Train on GTA5 (G)}\\
		\cline{6-10}
		&&&&& C&B&M&S&Mean\\
		\hline
		\multirow{5}{*}{ShuffleNetV2}&Baseline \cite{choi2021robustnet} &- &- &- &31.0 &32.1 &35.3 &24.3 &30.7 \\
		\cline{2-10}
		
		&\multirow{4}{1.0cm}{Ours}&$\surd$ &- &- &33.9 &32.5 &35.5 &28.1 &32.5  \\
		
		& &- &$\surd$ &- &34.8 &35.3 &35.9 &29.9  &34.0 \\
		
		& &- &- &$\surd$ &35.0 &34.1 &36.2 &29.8  &33.8 \\
		
		& &- &$\surd$ &$\surd$ &\textbf{35.4} &\textbf{35.9} &\textbf{36.3} &\textbf{30.9}  &\textbf{34.6} \\
		
		\hline
	\end{tabular}
	\label{tab:abl}
\end{table*}

\begin{figure*}[!t]
	\centering
	\includegraphics[width=1\textwidth,height=4.5cm]{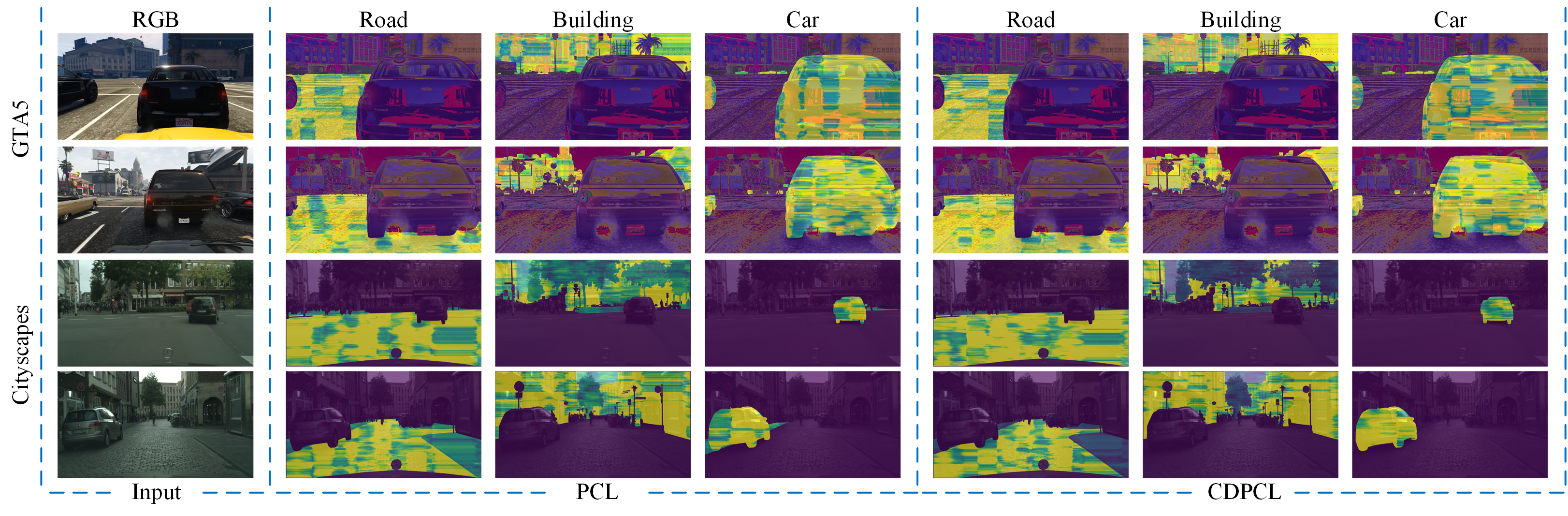}
	\caption{Activation maps visualization comparison of the weight matrix of the learned class-wise features between the PCL and our proposed approach (CDPCL) in the GTA5 to Cityscapes task.}
	\label{fig:vis_feats}
\end{figure*}

\subsection{Comparison with Multi-source Domain Generalization Methods}
To further verify the effectiveness of our proposed approach, we compare our proposed approach with recent approaches \cite{choi2021robustnet, li2018learning, zhao2022style, kim2022pin} under the multi-source domain generalization setting. The experimental results are shown in Table \ref{tab:ms_gs}. As shown in Table \ref{tab:ms_gs}, our proposed approach respectively achieves the improvement of 4.9\% and 1.6\% in average mIoU over the SOTA methods PinMem \cite{kim2022pin} and SHADE \cite{zhao2022style}. In conclusion, with richer source domains during the training process, our proposed approach can generate more prototypes of different domains and better learn class-wise domain-invariant features from the prototypes of different domains to improve the generalization ability for semantic segmentation. More experiments of multi-source domain generalization settings are shown in the Appendix \ref{app_exp_msdg}.

\subsection{Ablation Study}
\label{ablation}
In this section, four group experiments are conducted on the ``G$\rightarrow$\{C, B, M, S\}'' generalization setting to analyze the contributions of each component, including the prototypical contrastive learning (PCL), the uncertainty-weighted prototypical contrastive learning (UPCL), and the hard-weighted prototypical contrastive learning (HPCL), to the final performance and verify our motivation. 

The results are given in Table \ref{tab:abl}. We observe that the performance of the UPCL and HPCL respectively achieve an average improvement of 1.5\% and 1.3\% than the performance of the PCL. The performance of our proposed approach using the UPCL and HPCL simultaneously is superior to the performance of the PCL, UPCL, and HPCL. It demonstrates that the UPCL and HPCL can both boost the learning of class-wise domain-invariant features. Moreover, we visualize the weight matrix of the learned class-wise features on the source domain and the unseen domain. From Figure \ref{fig:vis_feats}, we can see that regions are activated by a weight matrix slot corresponding to each class. For the visualization of the PCL, some other classes are activated in the unseen domain (Cityscapes). For example, the class of ``sidewalk'' is activated in the activation maps of the ``road'' and ``car''. the classes of ``vegetation'' and ``sky'' are activated in the activation maps of the ``building''. These results demonstrate that there is still domain discrepancy between the learned class-wise features and the prototypes of the unseen domain. Compared with the visualization of the PCL, the visualization of our proposed approach (CDPCL) achieves a significant improvement. More qualitative and quantitative experiments about the discrepancy change between the learned class-wise features and the different domains before and after applying the proposed approach are analyzed which are shown in the Appendix \ref{app_exp_ddc}. In addition, the weights of the activated region corresponding to each class in our proposed approach is higher than the PCL. It demonstrates that our proposed approach can better learn class-wise domain-invariant features. More activation maps visualization and some visualization of the ablation study are shown in the Appendix \ref{app_sm}.

\begin{table}[htb]
	\centering
	{
		\caption{Computational complexity experiments based on the ResNet-50 backbone. Params denotes the number of parameters. FLOPs denotes the number of floating point operations.}
		\label{tab:time}
		\renewcommand\tabcolsep{3.0pt}
		\begin{tabular}{cccc}
			\hline
			Method& 
			FLOPs (G)& 
			Params (M)& 
			Inference time (ms) \\
			\hline
			baseline \cite{choi2021robustnet} &155.92&45.08&131.22  \\
			Ours &160.75&47.18&132.80 \\
			\hline
		\end{tabular}
	}
\end{table}

\subsection{Model complexity analysis}
\label{complexity}
We analyze the computational cost of our proposed approach. The input size is set as $1024 \times 512 \times 3$. The experiments are conducted on a single GTX 3090 GPU. As shown in Table \ref{tab:time}, for DeepLabV3+ with the ResNet-50 backbone, our proposed approach increases less than 5\% on parameters, floating point operations, and inference time. This shows that our proposed approach achieves a significant improvement in performance with very limited extra computational cost.

\section{Conclusion}
In this paper, a calibration-based dual prototypical contrastive learning (CDPCL) approach is proposed to reduce the domain discrepancy between the learned class-wise features and the prototypes of different domains for domain generalization semantic segmentation. The CDPCL approach contains an uncertainty-guided prototypical contrastive learning (UPCL) and a hard-weighted prototypical contrastive learning (HPCL). The proposed UPCL and HPCL respectively generate an uncertainty probability matrix and a hard-weighted matrix to calibrate the weights of the prototypes during the prototypical contrastive learning. Extensive experiments demonstrate that our approach achieves superior performance over current approaches on domain generalization semantic segmentation tasks.

\section*{Acknowledgment}
This work was supported in part by the National Natural Science Foundation of China under grants 62171294, 62101344, in part by the key Project of DEGP (Department of Education of Guangdong Province) under grants 2018KCXTD027, in part by the Natural Science Foundation of Guangdong Province, China under grants 2022A1515010159, 2020A1515010959, in part by the Key project of Shenzhen Science and Technology Plan under Grant 2022081018061\\7001, in part by the Interdisciplinary Innovation Team of Shenzhen University and in part by the Tencent ``Rhinoceros Birds'' - Scientific Research Foundation for Young Teachers of Shenzhen University, China.




\bibliographystyle{ACM-Reference-Format}
\balance
\bibliography{sample-base}


\begin{thebibliography}{54}


\ifx \showCODEN    \undefined \def \showCODEN     #1{\unskip}     \fi
\ifx \showDOI      \undefined \def \showDOI       #1{#1}\fi
\ifx \showISBNx    \undefined \def \showISBNx     #1{\unskip}     \fi
\ifx \showISBNxiii \undefined \def \showISBNxiii  #1{\unskip}     \fi
\ifx \showISSN     \undefined \def \showISSN      #1{\unskip}     \fi
\ifx \showLCCN     \undefined \def \showLCCN      #1{\unskip}     \fi
\ifx \shownote     \undefined \def \shownote      #1{#1}          \fi
\ifx \showarticletitle \undefined \def \showarticletitle #1{#1}   \fi
\ifx \showURL      \undefined \def \showURL       {\relax}        \fi
\providecommand\bibfield[2]{#2}
\providecommand\bibinfo[2]{#2}
\providecommand\natexlab[1]{#1}
\providecommand\showeprint[2][]{arXiv:#2}

\bibitem[Cai et~al\mbox{.}(2020)]%
        {cai2020generalizing}
\bibfield{author}{\bibinfo{person}{Minjie Cai}, \bibinfo{person}{Feng Lu},
  {and} \bibinfo{person}{Yoichi Sato}.} \bibinfo{year}{2020}\natexlab{}.
\newblock \showarticletitle{Generalizing hand segmentation in egocentric videos
  with uncertainty-guided model adaptation}. In
  \bibinfo{booktitle}{\emph{Proceedings of the ieee/cvf conference on computer
  vision and pattern recognition}}. \bibinfo{pages}{14392--14401}.
\newblock


\bibitem[Chen et~al\mbox{.}(2022)]%
        {chen2022compound}
\bibfield{author}{\bibinfo{person}{Chaoqi Chen}, \bibinfo{person}{Jiongcheng
  Li}, \bibinfo{person}{Xiaoguang Han}, \bibinfo{person}{Xiaoqing Liu}, {and}
  \bibinfo{person}{Yizhou Yu}.} \bibinfo{year}{2022}\natexlab{}.
\newblock \showarticletitle{Compound domain generalization via meta-knowledge
  encoding}. In \bibinfo{booktitle}{\emph{Proceedings of the IEEE/CVF
  Conference on Computer Vision and Pattern Recognition}}.
  \bibinfo{pages}{7119--7129}.
\newblock


\bibitem[Chen et~al\mbox{.}(2019)]%
        {chen2019progressive}
\bibfield{author}{\bibinfo{person}{Chaoqi Chen}, \bibinfo{person}{Weiping Xie},
  \bibinfo{person}{Wenbing Huang}, \bibinfo{person}{Yu Rong},
  \bibinfo{person}{Xinghao Ding}, \bibinfo{person}{Yue Huang},
  \bibinfo{person}{Tingyang Xu}, {and} \bibinfo{person}{Junzhou Huang}.}
  \bibinfo{year}{2019}\natexlab{}.
\newblock \showarticletitle{Progressive feature alignment for unsupervised
  domain adaptation}. In \bibinfo{booktitle}{\emph{Proceedings of the IEEE/CVF
  conference on computer vision and pattern recognition}}.
  \bibinfo{pages}{627--636}.
\newblock


\bibitem[Chen et~al\mbox{.}(2017)]%
        {chen2017deeplab}
\bibfield{author}{\bibinfo{person}{Liang-Chieh Chen}, \bibinfo{person}{George
  Papandreou}, \bibinfo{person}{Iasonas Kokkinos}, \bibinfo{person}{Kevin
  Murphy}, {and} \bibinfo{person}{Alan~L Yuille}.}
  \bibinfo{year}{2017}\natexlab{}.
\newblock \showarticletitle{Deeplab: Semantic image segmentation with deep
  convolutional nets, atrous convolution, and fully connected crfs}.
\newblock \bibinfo{journal}{\emph{IEEE transactions on pattern analysis and
  machine intelligence}} \bibinfo{volume}{40}, \bibinfo{number}{4}
  (\bibinfo{year}{2017}), \bibinfo{pages}{834--848}.
\newblock


\bibitem[Chen et~al\mbox{.}(2018)]%
        {chen2018encoder}
\bibfield{author}{\bibinfo{person}{Liang-Chieh Chen}, \bibinfo{person}{Yukun
  Zhu}, \bibinfo{person}{George Papandreou}, \bibinfo{person}{Florian Schroff},
  {and} \bibinfo{person}{Hartwig Adam}.} \bibinfo{year}{2018}\natexlab{}.
\newblock \showarticletitle{Encoder-decoder with atrous separable convolution
  for semantic image segmentation}. In \bibinfo{booktitle}{\emph{Proceedings of
  the European conference on computer vision (ECCV)}}.
  \bibinfo{pages}{801--818}.
\newblock


\bibitem[Choi et~al\mbox{.}(2021)]%
        {choi2021robustnet}
\bibfield{author}{\bibinfo{person}{Sungha Choi}, \bibinfo{person}{Sanghun
  Jung}, \bibinfo{person}{Huiwon Yun}, \bibinfo{person}{Joanne~T Kim},
  \bibinfo{person}{Seungryong Kim}, {and} \bibinfo{person}{Jaegul Choo}.}
  \bibinfo{year}{2021}\natexlab{}.
\newblock \showarticletitle{Robustnet: Improving domain generalization in
  urban-scene segmentation via instance selective whitening}. In
  \bibinfo{booktitle}{\emph{Proceedings of the IEEE/CVF Conference on Computer
  Vision and Pattern Recognition}}. \bibinfo{pages}{11580--11590}.
\newblock


\bibitem[Cordts et~al\mbox{.}(2016)]%
        {cordts2016cityscapes}
\bibfield{author}{\bibinfo{person}{Marius Cordts}, \bibinfo{person}{Mohamed
  Omran}, \bibinfo{person}{Sebastian Ramos}, \bibinfo{person}{Timo Rehfeld},
  \bibinfo{person}{Markus Enzweiler}, \bibinfo{person}{Rodrigo Benenson},
  \bibinfo{person}{Uwe Franke}, \bibinfo{person}{Stefan Roth}, {and}
  \bibinfo{person}{Bernt Schiele}.} \bibinfo{year}{2016}\natexlab{}.
\newblock \showarticletitle{The cityscapes dataset for semantic urban scene
  understanding}. In \bibinfo{booktitle}{\emph{Proceedings of the IEEE
  conference on computer vision and pattern recognition}}.
  \bibinfo{pages}{3213--3223}.
\newblock


\bibitem[He et~al\mbox{.}(2016)]%
        {he2016deep}
\bibfield{author}{\bibinfo{person}{Kaiming He}, \bibinfo{person}{Xiangyu
  Zhang}, \bibinfo{person}{Shaoqing Ren}, {and} \bibinfo{person}{Jian Sun}.}
  \bibinfo{year}{2016}\natexlab{}.
\newblock \showarticletitle{Deep residual learning for image recognition}. In
  \bibinfo{booktitle}{\emph{Proceedings of the IEEE conference on computer
  vision and pattern recognition}}. \bibinfo{pages}{770--778}.
\newblock


\bibitem[He et~al\mbox{.}(2021)]%
        {he2021fast}
\bibfield{author}{\bibinfo{person}{Yulin He}, \bibinfo{person}{Wei Chen},
  \bibinfo{person}{Zhengfa Liang}, \bibinfo{person}{Dan Chen},
  \bibinfo{person}{Yusong Tan}, \bibinfo{person}{Xin Luo},
  \bibinfo{person}{Chen Li}, {and} \bibinfo{person}{Yulan Guo}.}
  \bibinfo{year}{2021}\natexlab{}.
\newblock \showarticletitle{Fast and Accurate Lane Detection via Frequency
  Domain Learning}. In \bibinfo{booktitle}{\emph{Proceedings of the 29th ACM
  International Conference on Multimedia}}. \bibinfo{pages}{890--898}.
\newblock


\bibitem[Hua et~al\mbox{.}(2023)]%
        {hua2023multiple}
\bibfield{author}{\bibinfo{person}{Guoguang Hua}, \bibinfo{person}{Muxin Liao},
  \bibinfo{person}{Shishun Tian}, \bibinfo{person}{Yuhang Zhang}, {and}
  \bibinfo{person}{Wenbin Zou}.} \bibinfo{year}{2023}\natexlab{}.
\newblock \showarticletitle{Multiple Relational Learning Network for Joint
  Referring Expression Comprehension and Segmentation}.
\newblock \bibinfo{journal}{\emph{IEEE Transactions on Multimedia}}
  (\bibinfo{year}{2023}).
\newblock


\bibitem[Huang et~al\mbox{.}(2021)]%
        {huang2021fsdr}
\bibfield{author}{\bibinfo{person}{Jiaxing Huang}, \bibinfo{person}{Dayan
  Guan}, \bibinfo{person}{Aoran Xiao}, {and} \bibinfo{person}{Shijian Lu}.}
  \bibinfo{year}{2021}\natexlab{}.
\newblock \showarticletitle{Fsdr: Frequency space domain randomization for
  domain generalization}. In \bibinfo{booktitle}{\emph{Proceedings of the
  IEEE/CVF Conference on Computer Vision and Pattern Recognition}}.
  \bibinfo{pages}{6891--6902}.
\newblock


\bibitem[Jiang et~al\mbox{.}(2022)]%
        {jiang2022prototypical}
\bibfield{author}{\bibinfo{person}{Zhengkai Jiang}, \bibinfo{person}{Yuxi Li},
  \bibinfo{person}{Ceyuan Yang}, \bibinfo{person}{Peng Gao},
  \bibinfo{person}{Yabiao Wang}, \bibinfo{person}{Ying Tai}, {and}
  \bibinfo{person}{Chengjie Wang}.} \bibinfo{year}{2022}\natexlab{}.
\newblock \showarticletitle{Prototypical contrast adaptation for domain
  adaptive semantic segmentation}. In \bibinfo{booktitle}{\emph{European
  Conference on Computer Vision}}. Springer, \bibinfo{pages}{36--54}.
\newblock


\bibitem[Kim et~al\mbox{.}(2022)]%
        {kim2022pin}
\bibfield{author}{\bibinfo{person}{Jin Kim}, \bibinfo{person}{Jiyoung Lee},
  \bibinfo{person}{Jungin Park}, \bibinfo{person}{Dongbo Min}, {and}
  \bibinfo{person}{Kwanghoon Sohn}.} \bibinfo{year}{2022}\natexlab{}.
\newblock \showarticletitle{Pin the Memory: Learning to Generalize Semantic
  Segmentation}. In \bibinfo{booktitle}{\emph{Proceedings of the IEEE/CVF
  Conference on Computer Vision and Pattern Recognition}}.
  \bibinfo{pages}{4350--4360}.
\newblock


\bibitem[Lee et~al\mbox{.}(2022a)]%
        {lee2022bi}
\bibfield{author}{\bibinfo{person}{Geon Lee}, \bibinfo{person}{Chanho Eom},
  \bibinfo{person}{Wonkyung Lee}, \bibinfo{person}{Hyekang Park}, {and}
  \bibinfo{person}{Bumsub Ham}.} \bibinfo{year}{2022}\natexlab{a}.
\newblock \showarticletitle{Bi-directional Contrastive Learning for Domain
  Adaptive Semantic Segmentation}. In \bibinfo{booktitle}{\emph{European
  Conference on Computer Vision}}. Springer, \bibinfo{pages}{38--55}.
\newblock


\bibitem[Lee et~al\mbox{.}(2022b)]%
        {lee2022wildnet}
\bibfield{author}{\bibinfo{person}{Suhyeon Lee}, \bibinfo{person}{Hongje
  Seong}, \bibinfo{person}{Seongwon Lee}, {and} \bibinfo{person}{Euntai Kim}.}
  \bibinfo{year}{2022}\natexlab{b}.
\newblock \showarticletitle{WildNet: Learning Domain Generalized Semantic
  Segmentation from the Wild}. In \bibinfo{booktitle}{\emph{Proceedings of the
  IEEE/CVF Conference on Computer Vision and Pattern Recognition}}.
  \bibinfo{pages}{9936--9946}.
\newblock


\bibitem[Li et~al\mbox{.}(2018)]%
        {li2018learning}
\bibfield{author}{\bibinfo{person}{Da Li}, \bibinfo{person}{Yongxin Yang},
  \bibinfo{person}{Yi-Zhe Song}, {and} \bibinfo{person}{Timothy Hospedales}.}
  \bibinfo{year}{2018}\natexlab{}.
\newblock \showarticletitle{Learning to generalize: Meta-learning for domain
  generalization}. In \bibinfo{booktitle}{\emph{Proceedings of the AAAI
  conference on artificial intelligence}}, Vol.~\bibinfo{volume}{32}.
\newblock


\bibitem[Li et~al\mbox{.}(2022)]%
        {li2022cross}
\bibfield{author}{\bibinfo{person}{Miaoyu Li}, \bibinfo{person}{Yachao Zhang},
  \bibinfo{person}{Yuan Xie}, \bibinfo{person}{Zuodong Gao},
  \bibinfo{person}{Cuihua Li}, \bibinfo{person}{Zhizhong Zhang}, {and}
  \bibinfo{person}{Yanyun Qu}.} \bibinfo{year}{2022}\natexlab{}.
\newblock \showarticletitle{Cross-Domain and Cross-Modal Knowledge Distillation
  in Domain Adaptation for 3D Semantic Segmentation}. In
  \bibinfo{booktitle}{\emph{Proceedings of the 30th ACM International
  Conference on Multimedia}}. \bibinfo{pages}{3829--3837}.
\newblock


\bibitem[Liao et~al\mbox{.}(2022)]%
        {liao2022exploring}
\bibfield{author}{\bibinfo{person}{Muxin Liao}, \bibinfo{person}{Guoguang Hua},
  \bibinfo{person}{Shishun Tian}, \bibinfo{person}{Yuhang Zhang},
  \bibinfo{person}{Wenbin Zou}, {and} \bibinfo{person}{Xia Li}.}
  \bibinfo{year}{2022}\natexlab{}.
\newblock \showarticletitle{Exploring More Concentrated and Consistent
  Activation Regions for Cross-domain Semantic Segmentation}.
\newblock \bibinfo{journal}{\emph{Neurocomputing}} (\bibinfo{year}{2022}).
\newblock


\bibitem[Lu et~al\mbox{.}(2022)]%
        {lu2022bidirectional}
\bibfield{author}{\bibinfo{person}{Yulei Lu}, \bibinfo{person}{Yawei Luo},
  \bibinfo{person}{Li Zhang}, \bibinfo{person}{Zheyang Li}, \bibinfo{person}{Yi
  Yang}, {and} \bibinfo{person}{Jun Xiao}.} \bibinfo{year}{2022}\natexlab{}.
\newblock \showarticletitle{Bidirectional self-training with multiple
  anisotropic prototypes for domain adaptive semantic segmentation}. In
  \bibinfo{booktitle}{\emph{Proceedings of the 30th ACM International
  Conference on Multimedia}}. \bibinfo{pages}{1405--1415}.
\newblock


\bibitem[Ma et~al\mbox{.}(2018)]%
        {ma2018shufflenet}
\bibfield{author}{\bibinfo{person}{Ningning Ma}, \bibinfo{person}{Xiangyu
  Zhang}, \bibinfo{person}{Hai-Tao Zheng}, {and} \bibinfo{person}{Jian Sun}.}
  \bibinfo{year}{2018}\natexlab{}.
\newblock \showarticletitle{Shufflenet v2: Practical guidelines for efficient
  cnn architecture design}. In \bibinfo{booktitle}{\emph{Proceedings of the
  European conference on computer vision (ECCV)}}. \bibinfo{pages}{116--131}.
\newblock


\bibitem[Ma et~al\mbox{.}(2022)]%
        {ma2022rethinking}
\bibfield{author}{\bibinfo{person}{Zeyu Ma}, \bibinfo{person}{Yang Yang},
  \bibinfo{person}{Guoqing Wang}, \bibinfo{person}{Xing Xu},
  \bibinfo{person}{Heng~Tao Shen}, {and} \bibinfo{person}{Mingxing Zhang}.}
  \bibinfo{year}{2022}\natexlab{}.
\newblock \showarticletitle{Rethinking Open-World Object Detection in
  Autonomous Driving Scenarios}. In \bibinfo{booktitle}{\emph{Proceedings of
  the 30th ACM International Conference on Multimedia}}.
  \bibinfo{pages}{1279--1288}.
\newblock


\bibitem[Neuhold et~al\mbox{.}(2017)]%
        {neuhold2017mapillary}
\bibfield{author}{\bibinfo{person}{Gerhard Neuhold}, \bibinfo{person}{Tobias
  Ollmann}, \bibinfo{person}{Samuel Rota~Bulo}, {and} \bibinfo{person}{Peter
  Kontschieder}.} \bibinfo{year}{2017}\natexlab{}.
\newblock \showarticletitle{The mapillary vistas dataset for semantic
  understanding of street scenes}. In \bibinfo{booktitle}{\emph{Proceedings of
  the IEEE international conference on computer vision}}.
  \bibinfo{pages}{4990--4999}.
\newblock


\bibitem[Pan et~al\mbox{.}(2018)]%
        {pan2018two}
\bibfield{author}{\bibinfo{person}{Xingang Pan}, \bibinfo{person}{Ping Luo},
  \bibinfo{person}{Jianping Shi}, {and} \bibinfo{person}{Xiaoou Tang}.}
  \bibinfo{year}{2018}\natexlab{}.
\newblock \showarticletitle{Two at once: Enhancing learning and generalization
  capacities via ibn-net}. In \bibinfo{booktitle}{\emph{Proceedings of the
  European Conference on Computer Vision (ECCV)}}. \bibinfo{pages}{464--479}.
\newblock


\bibitem[Pan et~al\mbox{.}(2019)]%
        {pan2019switchable}
\bibfield{author}{\bibinfo{person}{Xingang Pan}, \bibinfo{person}{Xiaohang
  Zhan}, \bibinfo{person}{Jianping Shi}, \bibinfo{person}{Xiaoou Tang}, {and}
  \bibinfo{person}{Ping Luo}.} \bibinfo{year}{2019}\natexlab{}.
\newblock \showarticletitle{Switchable whitening for deep representation
  learning}. In \bibinfo{booktitle}{\emph{Proceedings of the IEEE/CVF
  International Conference on Computer Vision}}. \bibinfo{pages}{1863--1871}.
\newblock


\bibitem[Peng et~al\mbox{.}(2022a)]%
        {peng2022semantic}
\bibfield{author}{\bibinfo{person}{Duo Peng}, \bibinfo{person}{Yinjie Lei},
  \bibinfo{person}{Munawar Hayat}, \bibinfo{person}{Yulan Guo}, {and}
  \bibinfo{person}{Wen Li}.} \bibinfo{year}{2022}\natexlab{a}.
\newblock \showarticletitle{Semantic-aware domain generalized segmentation}. In
  \bibinfo{booktitle}{\emph{Proceedings of the IEEE/CVF Conference on Computer
  Vision and Pattern Recognition}}. \bibinfo{pages}{2594--2605}.
\newblock


\bibitem[Peng et~al\mbox{.}(2021)]%
        {peng2021global}
\bibfield{author}{\bibinfo{person}{Duo Peng}, \bibinfo{person}{Yinjie Lei},
  \bibinfo{person}{Lingqiao Liu}, \bibinfo{person}{Pingping Zhang}, {and}
  \bibinfo{person}{Jun Liu}.} \bibinfo{year}{2021}\natexlab{}.
\newblock \showarticletitle{Global and local texture randomization for
  synthetic-to-real semantic segmentation}.
\newblock \bibinfo{journal}{\emph{IEEE Transactions on Image Processing}}
  \bibinfo{volume}{30} (\bibinfo{year}{2021}), \bibinfo{pages}{6594--6608}.
\newblock


\bibitem[Peng et~al\mbox{.}(2022b)]%
        {peng2022out}
\bibfield{author}{\bibinfo{person}{Xi Peng}, \bibinfo{person}{Fengchun Qiao},
  {and} \bibinfo{person}{Long Zhao}.} \bibinfo{year}{2022}\natexlab{b}.
\newblock \showarticletitle{Out-of-domain generalization from a single source:
  An uncertainty quantification approach}.
\newblock \bibinfo{journal}{\emph{IEEE Transactions on Pattern Analysis and
  Machine Intelligence}} (\bibinfo{year}{2022}).
\newblock


\bibitem[Qiao and Peng(2021)]%
        {qiao2021uncertainty}
\bibfield{author}{\bibinfo{person}{Fengchun Qiao} {and} \bibinfo{person}{Xi
  Peng}.} \bibinfo{year}{2021}\natexlab{}.
\newblock \showarticletitle{Uncertainty-guided model generalization to unseen
  domains}. In \bibinfo{booktitle}{\emph{Proceedings of the IEEE/CVF conference
  on computer vision and pattern recognition}}. \bibinfo{pages}{6790--6800}.
\newblock


\bibitem[Richter et~al\mbox{.}(2016)]%
        {richter2016playing}
\bibfield{author}{\bibinfo{person}{Stephan~R Richter}, \bibinfo{person}{Vibhav
  Vineet}, \bibinfo{person}{Stefan Roth}, {and} \bibinfo{person}{Vladlen
  Koltun}.} \bibinfo{year}{2016}\natexlab{}.
\newblock \showarticletitle{Playing for data: Ground truth from computer
  games}. In \bibinfo{booktitle}{\emph{European conference on computer
  vision}}. Springer, \bibinfo{pages}{102--118}.
\newblock


\bibitem[Ros et~al\mbox{.}(2016)]%
        {ros2016synthia}
\bibfield{author}{\bibinfo{person}{German Ros}, \bibinfo{person}{Laura
  Sellart}, \bibinfo{person}{Joanna Materzynska}, \bibinfo{person}{David
  Vazquez}, {and} \bibinfo{person}{Antonio~M Lopez}.}
  \bibinfo{year}{2016}\natexlab{}.
\newblock \showarticletitle{The synthia dataset: A large collection of
  synthetic images for semantic segmentation of urban scenes}. In
  \bibinfo{booktitle}{\emph{Proceedings of the IEEE conference on computer
  vision and pattern recognition}}. \bibinfo{pages}{3234--3243}.
\newblock


\bibitem[Sandler et~al\mbox{.}(2018)]%
        {sandler2018mobilenetv2}
\bibfield{author}{\bibinfo{person}{Mark Sandler}, \bibinfo{person}{Andrew
  Howard}, \bibinfo{person}{Menglong Zhu}, \bibinfo{person}{Andrey Zhmoginov},
  {and} \bibinfo{person}{Liang-Chieh Chen}.} \bibinfo{year}{2018}\natexlab{}.
\newblock \showarticletitle{Mobilenetv2: Inverted residuals and linear
  bottlenecks}. In \bibinfo{booktitle}{\emph{Proceedings of the IEEE conference
  on computer vision and pattern recognition}}. \bibinfo{pages}{4510--4520}.
\newblock


\bibitem[Su et~al\mbox{.}(2022)]%
        {su2022consistency}
\bibfield{author}{\bibinfo{person}{Siwei Su}, \bibinfo{person}{Haijian Wang},
  {and} \bibinfo{person}{Meng Yang}.} \bibinfo{year}{2022}\natexlab{}.
\newblock \showarticletitle{Consistency Learning based on Class-Aware Style
  Variation for Domain Generalizable Semantic Segmentation}. In
  \bibinfo{booktitle}{\emph{Proceedings of the 30th ACM International
  Conference on Multimedia}}. \bibinfo{pages}{6029--6038}.
\newblock


\bibitem[Tjio et~al\mbox{.}(2022)]%
        {tjio2022adversarial}
\bibfield{author}{\bibinfo{person}{Gabriel Tjio}, \bibinfo{person}{Ping Liu},
  \bibinfo{person}{Joey~Tianyi Zhou}, {and} \bibinfo{person}{Rick Siow~Mong
  Goh}.} \bibinfo{year}{2022}\natexlab{}.
\newblock \showarticletitle{Adversarial semantic hallucination for domain
  generalized semantic segmentation}. In \bibinfo{booktitle}{\emph{Proceedings
  of the IEEE/CVF Winter Conference on Applications of Computer Vision}}.
  \bibinfo{pages}{318--327}.
\newblock


\bibitem[Varma et~al\mbox{.}(2019)]%
        {varma2019idd}
\bibfield{author}{\bibinfo{person}{Girish Varma}, \bibinfo{person}{Anbumani
  Subramanian}, \bibinfo{person}{Anoop Namboodiri}, \bibinfo{person}{Manmohan
  Chandraker}, {and} \bibinfo{person}{CV Jawahar}.}
  \bibinfo{year}{2019}\natexlab{}.
\newblock \showarticletitle{IDD: A dataset for exploring problems of autonomous
  navigation in unconstrained environments}. In \bibinfo{booktitle}{\emph{2019
  IEEE Winter Conference on Applications of Computer Vision (WACV)}}. IEEE,
  \bibinfo{pages}{1743--1751}.
\newblock


\bibitem[Wang et~al\mbox{.}(2022a)]%
        {wang2022domain}
\bibfield{author}{\bibinfo{person}{Jingye Wang}, \bibinfo{person}{Ruoyi Du},
  \bibinfo{person}{Dongliang Chang}, \bibinfo{person}{Kongming Liang}, {and}
  \bibinfo{person}{Zhanyu Ma}.} \bibinfo{year}{2022}\natexlab{a}.
\newblock \showarticletitle{Domain Generalization via Frequency-domain-based
  Feature Disentanglement and Interaction}. In
  \bibinfo{booktitle}{\emph{Proceedings of the 30th ACM International
  Conference on Multimedia}}. \bibinfo{pages}{4821--4829}.
\newblock


\bibitem[Wang et~al\mbox{.}(2023)]%
        {wang2023bp}
\bibfield{author}{\bibinfo{person}{Shanshan Wang}, \bibinfo{person}{Lei Zhang},
  \bibinfo{person}{Pichao Wang}, \bibinfo{person}{MengZhu Wang}, {and}
  \bibinfo{person}{Xingyi Zhang}.} \bibinfo{year}{2023}\natexlab{}.
\newblock \showarticletitle{BP-triplet net for unsupervised domain adaptation:
  A Bayesian perspective}.
\newblock \bibinfo{journal}{\emph{Pattern Recognition}}  \bibinfo{volume}{133}
  (\bibinfo{year}{2023}), \bibinfo{pages}{108993}.
\newblock


\bibitem[Wang et~al\mbox{.}(2022b)]%
        {wang2022feature}
\bibfield{author}{\bibinfo{person}{Yue Wang}, \bibinfo{person}{Lei Qi},
  \bibinfo{person}{Yinghuan Shi}, {and} \bibinfo{person}{Yang Gao}.}
  \bibinfo{year}{2022}\natexlab{b}.
\newblock \showarticletitle{Feature-based Style Randomization for Domain
  Generalization}.
\newblock \bibinfo{journal}{\emph{IEEE Transactions on Circuits and Systems for
  Video Technology}} (\bibinfo{year}{2022}).
\newblock


\bibitem[Wang et~al\mbox{.}(2020)]%
        {wang2020learning}
\bibfield{author}{\bibinfo{person}{Yinduo Wang}, \bibinfo{person}{Haofeng
  Zhang}, \bibinfo{person}{Zheng Zhang}, \bibinfo{person}{Yang Long}, {and}
  \bibinfo{person}{Ling Shao}.} \bibinfo{year}{2020}\natexlab{}.
\newblock \showarticletitle{Learning discriminative domain-invariant prototypes
  for generalized zero shot learning}.
\newblock \bibinfo{journal}{\emph{Knowledge-Based Systems}}
  \bibinfo{volume}{196} (\bibinfo{year}{2020}), \bibinfo{pages}{105796}.
\newblock


\bibitem[Wei et~al\mbox{.}(2022)]%
        {wei2022sginet}
\bibfield{author}{\bibinfo{person}{Yanyan Wei}, \bibinfo{person}{Zhao Zhang},
  \bibinfo{person}{Huan Zheng}, \bibinfo{person}{Richang Hong},
  \bibinfo{person}{Yi Yang}, {and} \bibinfo{person}{Meng Wang}.}
  \bibinfo{year}{2022}\natexlab{}.
\newblock \showarticletitle{Sginet: Toward sufficient interaction between
  single image deraining and semantic segmentation}. In
  \bibinfo{booktitle}{\emph{Proceedings of the 30th ACM International
  Conference on Multimedia}}. \bibinfo{pages}{6202--6210}.
\newblock


\bibitem[Xiao et~al\mbox{.}(2021)]%
        {xiao2021bit}
\bibfield{author}{\bibinfo{person}{Zehao Xiao}, \bibinfo{person}{Jiayi Shen},
  \bibinfo{person}{Xiantong Zhen}, \bibinfo{person}{Ling Shao}, {and}
  \bibinfo{person}{Cees Snoek}.} \bibinfo{year}{2021}\natexlab{}.
\newblock \showarticletitle{A bit more bayesian: Domain-invariant learning with
  uncertainty}. In \bibinfo{booktitle}{\emph{International Conference on
  Machine Learning}}. PMLR, \bibinfo{pages}{11351--11361}.
\newblock


\bibitem[Xu et~al\mbox{.}(2022)]%
        {xu2022dirl}
\bibfield{author}{\bibinfo{person}{Qi Xu}, \bibinfo{person}{Liang Yao},
  \bibinfo{person}{Zhengkai Jiang}, \bibinfo{person}{Guannan Jiang},
  \bibinfo{person}{Wenqing Chu}, \bibinfo{person}{Wenhui Han},
  \bibinfo{person}{Wei Zhang}, \bibinfo{person}{Chengjie Wang}, {and}
  \bibinfo{person}{Ying Tai}.} \bibinfo{year}{2022}\natexlab{}.
\newblock \showarticletitle{DIRL: Domain-invariant representation learning for
  generalizable semantic segmentation}. In
  \bibinfo{booktitle}{\emph{Proceedings of the AAAI Conference on Artificial
  Intelligence}}, Vol.~\bibinfo{volume}{36}. \bibinfo{pages}{2884--2892}.
\newblock


\bibitem[Ye et~al\mbox{.}(2022)]%
        {ye2022alleviating}
\bibfield{author}{\bibinfo{person}{Yalan Ye}, \bibinfo{person}{Ziqi Liu},
  \bibinfo{person}{Yangwuyong Zhang}, \bibinfo{person}{Jingjing Li}, {and}
  \bibinfo{person}{Hengtao Shen}.} \bibinfo{year}{2022}\natexlab{}.
\newblock \showarticletitle{Alleviating Style Sensitivity then Adapting:
  Source-free Domain Adaptation for Medical Image Segmentation}. In
  \bibinfo{booktitle}{\emph{Proceedings of the 30th ACM International
  Conference on Multimedia}}. \bibinfo{pages}{1935--1944}.
\newblock


\bibitem[Yu et~al\mbox{.}(2020)]%
        {yu2020bdd100k}
\bibfield{author}{\bibinfo{person}{Fisher Yu}, \bibinfo{person}{Haofeng Chen},
  \bibinfo{person}{Xin Wang}, \bibinfo{person}{Wenqi Xian},
  \bibinfo{person}{Yingying Chen}, \bibinfo{person}{Fangchen Liu},
  \bibinfo{person}{Vashisht Madhavan}, {and} \bibinfo{person}{Trevor Darrell}.}
  \bibinfo{year}{2020}\natexlab{}.
\newblock \showarticletitle{Bdd100k: A diverse driving dataset for
  heterogeneous multitask learning}. In \bibinfo{booktitle}{\emph{Proceedings
  of the IEEE/CVF conference on computer vision and pattern recognition}}.
  \bibinfo{pages}{2636--2645}.
\newblock


\bibitem[Yu et~al\mbox{.}(2022)]%
        {yu2022mttrans}
\bibfield{author}{\bibinfo{person}{Jinze Yu}, \bibinfo{person}{Jiaming Liu},
  \bibinfo{person}{Xiaobao Wei}, \bibinfo{person}{Haoyi Zhou},
  \bibinfo{person}{Yohei Nakata}, \bibinfo{person}{Denis Gudovskiy},
  \bibinfo{person}{Tomoyuki Okuno}, \bibinfo{person}{Jianxin Li},
  \bibinfo{person}{Kurt Keutzer}, {and} \bibinfo{person}{Shanghang Zhang}.}
  \bibinfo{year}{2022}\natexlab{}.
\newblock \showarticletitle{MTTrans: Cross-domain Object Detection with Mean
  Teacher Transformer}. In \bibinfo{booktitle}{\emph{Computer Vision--ECCV
  2022: 17th European Conference, Tel Aviv, Israel, October 23--27, 2022,
  Proceedings, Part IX}}. Springer, \bibinfo{pages}{629--645}.
\newblock


\bibitem[Yuan et~al\mbox{.}(2023)]%
        {yuan2023prototype}
\bibfield{author}{\bibinfo{person}{Zhimin Yuan}, \bibinfo{person}{Ming Cheng},
  \bibinfo{person}{Wankang Zeng}, \bibinfo{person}{Yanfei Su},
  \bibinfo{person}{Weiquan Liu}, \bibinfo{person}{Shangshu Yu}, {and}
  \bibinfo{person}{Cheng Wang}.} \bibinfo{year}{2023}\natexlab{}.
\newblock \showarticletitle{Prototype-guided Multi-task Adversarial Network for
  Cross-domain LiDAR Point Clouds Semantic Segmentation}.
\newblock \bibinfo{journal}{\emph{IEEE Transactions on Geoscience and Remote
  Sensing}} (\bibinfo{year}{2023}).
\newblock


\bibitem[Yue et~al\mbox{.}(2019)]%
        {yue2019domain}
\bibfield{author}{\bibinfo{person}{Xiangyu Yue}, \bibinfo{person}{Yang Zhang},
  \bibinfo{person}{Sicheng Zhao}, \bibinfo{person}{Alberto
  Sangiovanni-Vincentelli}, \bibinfo{person}{Kurt Keutzer}, {and}
  \bibinfo{person}{Boqing Gong}.} \bibinfo{year}{2019}\natexlab{}.
\newblock \showarticletitle{Domain randomization and pyramid consistency:
  Simulation-to-real generalization without accessing target domain data}. In
  \bibinfo{booktitle}{\emph{Proceedings of the IEEE/CVF International
  Conference on Computer Vision}}. \bibinfo{pages}{2100--2110}.
\newblock


\bibitem[Zhang et~al\mbox{.}(2022b)]%
        {zhang2022generalizable}
\bibfield{author}{\bibinfo{person}{Jian Zhang}, \bibinfo{person}{Lei Qi},
  \bibinfo{person}{Yinghuan Shi}, {and} \bibinfo{person}{Yang Gao}.}
  \bibinfo{year}{2022}\natexlab{b}.
\newblock \showarticletitle{Generalizable model-agnostic semantic segmentation
  via target-specific normalization}.
\newblock \bibinfo{journal}{\emph{Pattern Recognition}}  \bibinfo{volume}{122}
  (\bibinfo{year}{2022}), \bibinfo{pages}{108292}.
\newblock


\bibitem[Zhang et~al\mbox{.}(2022c)]%
        {zhang2022probabilistic}
\bibfield{author}{\bibinfo{person}{Wei Zhang}, \bibinfo{person}{Xiaohong
  Zhang}, \bibinfo{person}{Sheng Huang}, \bibinfo{person}{Yuting Lu}, {and}
  \bibinfo{person}{Kun Wang}.} \bibinfo{year}{2022}\natexlab{c}.
\newblock \showarticletitle{A Probabilistic Model for Controlling Diversity and
  Accuracy of Ambiguous Medical Image Segmentation}. In
  \bibinfo{booktitle}{\emph{Proceedings of the 30th ACM International
  Conference on Multimedia}}. \bibinfo{pages}{4751--4759}.
\newblock


\bibitem[Zhang et~al\mbox{.}(2022a)]%
        {zhang2022self}
\bibfield{author}{\bibinfo{person}{Yachao Zhang}, \bibinfo{person}{Miaoyu Li},
  \bibinfo{person}{Yuan Xie}, \bibinfo{person}{Cuihua Li},
  \bibinfo{person}{Cong Wang}, \bibinfo{person}{Zhizhong Zhang}, {and}
  \bibinfo{person}{Yanyun Qu}.} \bibinfo{year}{2022}\natexlab{a}.
\newblock \showarticletitle{Self-supervised Exclusive Learning for 3D
  Segmentation with Cross-Modal Unsupervised Domain Adaptation}. In
  \bibinfo{booktitle}{\emph{Proceedings of the 30th ACM International
  Conference on Multimedia}}. \bibinfo{pages}{3338--3346}.
\newblock


\bibitem[Zhang et~al\mbox{.}(2023)]%
        {zhang2023hybrid}
\bibfield{author}{\bibinfo{person}{Yuhang Zhang}, \bibinfo{person}{Shishun
  Tian}, \bibinfo{person}{Muxin Liao}, \bibinfo{person}{Wenbin Zou}, {and}
  \bibinfo{person}{Chen Xu}.} \bibinfo{year}{2023}\natexlab{}.
\newblock \showarticletitle{A hybrid domain learning framework for unsupervised
  semantic segmentation}.
\newblock \bibinfo{journal}{\emph{Neurocomputing}}  \bibinfo{volume}{516}
  (\bibinfo{year}{2023}), \bibinfo{pages}{133--145}.
\newblock


\bibitem[Zhang et~al\mbox{.}(2021)]%
        {zhang2021rpn}
\bibfield{author}{\bibinfo{person}{Yixin Zhang}, \bibinfo{person}{Zilei Wang},
  {and} \bibinfo{person}{Yushi Mao}.} \bibinfo{year}{2021}\natexlab{}.
\newblock \showarticletitle{Rpn prototype alignment for domain adaptive object
  detector}. In \bibinfo{booktitle}{\emph{Proceedings of the IEEE/CVF
  conference on computer vision and pattern recognition}}.
  \bibinfo{pages}{12425--12434}.
\newblock


\bibitem[Zhao et~al\mbox{.}(2022)]%
        {zhao2022style}
\bibfield{author}{\bibinfo{person}{Yuyang Zhao}, \bibinfo{person}{Zhun Zhong},
  \bibinfo{person}{Na Zhao}, \bibinfo{person}{Nicu Sebe}, {and}
  \bibinfo{person}{Gim~Hee Lee}.} \bibinfo{year}{2022}\natexlab{}.
\newblock \showarticletitle{Style-hallucinated dual consistency learning for
  domain generalized semantic segmentation}. In
  \bibinfo{booktitle}{\emph{Computer Vision--ECCV 2022: 17th European
  Conference, Tel Aviv, Israel, October 23--27, 2022, Proceedings, Part
  XXVIII}}. Springer, \bibinfo{pages}{535--552}.
\newblock


\bibitem[Zhu et~al\mbox{.}(2017)]%
        {zhu2017unpaired}
\bibfield{author}{\bibinfo{person}{Jun-Yan Zhu}, \bibinfo{person}{Taesung
  Park}, \bibinfo{person}{Phillip Isola}, {and} \bibinfo{person}{Alexei~A
  Efros}.} \bibinfo{year}{2017}\natexlab{}.
\newblock \showarticletitle{Unpaired image-to-image translation using
  cycle-consistent adversarial networks}. In
  \bibinfo{booktitle}{\emph{Proceedings of the IEEE international conference on
  computer vision}}. \bibinfo{pages}{2223--2232}.
\newblock


\bibitem[Zou et~al\mbox{.}(2023)]%
        {zou2023dual}
\bibfield{author}{\bibinfo{person}{Wenbin Zou}, \bibinfo{person}{Ruijing Long},
  \bibinfo{person}{Yuhang Zhang}, \bibinfo{person}{Muxin Liao},
  \bibinfo{person}{Zhi Zhou}, {and} \bibinfo{person}{Shishun Tian}.}
  \bibinfo{year}{2023}\natexlab{}.
\newblock \showarticletitle{Dual geometric perception for cross-domain road
  segmentation}.
\newblock \bibinfo{journal}{\emph{Displays}}  \bibinfo{volume}{76}
  (\bibinfo{year}{2023}), \bibinfo{pages}{102332}.
\newblock


\end{thebibliography}

\clearpage
\appendix

\section{Experiments}
\subsection{Datasets} 
\label{app_exp_data}
Our approach is evaluated on five standard single-source benchmarks and a standard multi-source benchmark. Five standard single-source benchmarks contain ``G$\rightarrow$\{S, C, M, B\}'', ``S$\rightarrow$\{G, C, M, B\}'', ``C$\rightarrow$\{G, S, M, B\}'', ``B$\rightarrow$\{G, S, C, M\}'', and ``M$\rightarrow$\{G, S, C, B\}''. The two standard multi-source benchmarks are ``\{G, S\}$\rightarrow$\{C, B, M\}'' and ``\{G, S, I\}$\rightarrow$\{C, B, M\}". The ``G'' and ``S'' mean GTA5 \cite{richter2016playing} and SYNTHIA \cite{ros2016synthia} datasets which are two synthetic datasets. The ``C'', ``B'', ``M'', and ``I'' mean the Cityscapes \cite{cordts2016cityscapes}, BDD \cite{yu2020bdd100k}, Mapillary \cite{neuhold2017mapillary}, and IDD \cite{varma2019idd} datasets which are three real-world datasets.
\subsection{Implementation Details}
\label{app_exp_id}
In our experiments, the ISW \cite{choi2021robustnet} is adopted as baseline. The ResNet-50 \cite{he2016deep}, ShuffleNetV2 \cite{ma2018shufflenet}, and MobileNetV2 \cite{sandler2018mobilenetv2} are utilized in DeepLabV3+ \cite{chen2018encoder} as the segmentation network. We follow the data augmentations of previous works, including ISW \cite{choi2021robustnet}, SAN \cite{peng2022semantic}, WildNet \cite{lee2022wildnet}, PinMemory \cite{kim2022pin}, and SHADE \cite{zhao2022style}. Specifically, color jittering (brightness of $0.4$, contrast of $0.4$, saturation of $0.4$, and hue of $0.1$), Gaussian blur, random cropping, random horizontal flipping, and random scaling with the range of $[0.5, 2.0]$ are used in our approach. The input images of five datasets are cropped to the resolution of $768 \times 768$. The mean Intersection-Over-Union value ($mIoU=\frac{TP}{TP+FP+FN}$) is utilized as the metric of evaluation, where TP, FP, and FN are denoted as the predicted pixels numbers of true positive, false positive, and false negative. The Stochastic Gradient Decent (SGD) optimizer with an initial learning rate of $1e-2$ and a momentum of 0.9 is leveraged to optimize our backbone network. In the training stages, the learning rate is adjusted by the power of 0.9 according to the polynomial learning rate scheduler and the maximum number of iterations is set to 40k steps. Moreover, the weight coefficients $m_p$, $m_a$, $m_u$, $\tau_u$, $\tau_h$, $\lambda_1$, and $\lambda_2$ are set as 0.9, 0.9, 0.9, 0.8, 0.8, 0.1, and 0.01 for all experiments.

\begin{table}[!t]
	\centering
	{
		\caption{Performance comparison in terms of mIoU (\%) between domain generalization methods in the architecture of the ResNet-50 \cite{he2016deep}. The best results are marked in bold and the second-best results are underlined.}
		\label{tab:ms_gsi}
		\renewcommand\tabcolsep{2.0pt}
		\begin{tabular}{lccccc}
			\hline
			\multirow{2}{*}{Methods}& 
			\multirow{2}{*}{Backbone}& 
			\multirow{2}{*}{Mean}& 
			\multicolumn{3}{c}{Train on G, S, and I}
			\\
			\cline{4-6}
			&&&  $\rightarrow$C&$\rightarrow$B&$\rightarrow$M \\
			\hline
			
			IBN \cite{pan2018two} & \multirow{6}{*}{ResNet-50} &53.1 &54.4 &48.9 &56.1 \\
			
			MLDG \cite{li2018learning} & &53.1 &54.8 &48.5 &55.9 \\
			
			ISW \cite{choi2021robustnet} & &53.5 &54.7 &49.0 &56.9 \\
			
			TSMLDG \cite{zhang2022generalizable} & &50.7 &53.0 &46.4 &52.8 \\
			
			PinMem \cite{kim2022pin} & &\underline{55.0} &\underline{56.6} &\underline{50.2} &\underline{58.3} \\
			
			Ours& &\textbf{59.5} &\textbf{61.1} &\textbf{54.8} &\textbf{62.5} \\
			\hline
		\end{tabular}
	}
\end{table}

\subsection{Comparison with Multi-source Domain Generalization Methods}
\label{app_exp_msdg}
Our proposed approach is trained on three source domains to compare with recent methods, including IBN \cite{pan2018two}, MLDG \cite{li2018learning}, ISW \cite{choi2021robustnet}, TSMLDG \cite{zhang2022generalizable}, and PinMem \cite{kim2022pin}, for further verifying the effectiveness of our proposed approach. As shown in Table \ref{tab:ms_gsi}, our proposed approach respectively achieves an improvement of 4.5\% in average mIoU over the PinMem \cite{kim2022pin}. Thus, with richer source domains during the training process, our proposed approach can better learn class-wise domain-invariant features from the prototypes of different domains to improve the generalization ability for semantic segmentation.

\begin{table}[htb]
	\centering
	{
		\caption{The discrepancy between the learned class-wise features and the prototypes of the source domain.}
		\label{tab:dc1}
		\renewcommand\tabcolsep{2.0pt}
		\begin{tabular}{ccc}
			\hline
			Class& 
			PCL& 
			CDPCL
			\\
			\hline
			Road & 0.0406 & 0.0337 \\
			Sidewalk & 0.0489 & 0.0504 \\
			Building & 0.0516 & 0.0493 \\
			Car & 0.0331 & 0.0289 \\
			\hline
		\end{tabular}
	}
\end{table}

\begin{table}[htb]
	\centering
	{
		\caption{The discrepancy between the learned class-wise features and the prototypes of the augmented domain.}
		\label{tab:dc2}
		\renewcommand\tabcolsep{2.0pt}
		\begin{tabular}{ccc}
			\hline
			Class& 
			PCL& 
			CDPCL
			\\
			\hline
			Road & 0.0571 & 0.0458 \\
			Sidewalk & 0.0494 & 0.0383 \\
			Building & 0.0508 & 0.0462 \\
			Car & 0.0414 & 0.0364 \\
			\hline
		\end{tabular}
	}
\end{table}

\subsection{The Analysis of The Distribution Discrepancy Change}
\label{app_exp_ddc}
We analyze the distribution discrepancy change between the learned class-wise features and the different domains before and after applying the proposed approach from two aspects, including qualitative and quantitative experiments.

\textbf{Qualitative Results.} We visualize activation maps of the weight matrix of the learned class-wise features between the conventional prototypical contrastive learning (PCL)  and the proposed approach (CDPCL), which are shown in Figure \ref{fig:vis_feats}. In the visualization of the PCL, some other classes are activated in the unseen domain (Cityscapes). For example, the classes of ''vegetation'' and ''sky'' are activated in the activation maps of the ''building''. These results demonstrate that there is still domain discrepancy between the learned class-wise features and the prototypes of the unseen domain (Cityscapes). Compared with the visualization of the PCL, the visualization of our proposed approach (CDPCL) achieves a significant improvement. It demonstrates that the difference between the learned class-wise features and the prototypes of the unseen domain (Cityscapes) is reduced.

\begin{table*}[!t]
	\centering
	{
		\caption{The similarity of different classes between the learned class-wise features and the source domain prototypes. The first row represents the learned class-wise features and the first column represents the source domain prototypes.}
		\label{tab:dc3}
		\renewcommand\tabcolsep{2.0pt}
		\begin{tabular}{cccccc|ccccc}
			\hline
			& 
			\multicolumn{4}{c}{PCL}& 
			&
			&
			\multicolumn{4}{c}{CDPCL}
			\\
			\hline
			Class & Road & Sidewalk & Building & Car & & Class & Road & Sidewalk & Building & Car\\
			Road & 0.8598 & 0.5131 & 0.4979 & 0.5368 & & Road & 0.9182 & 0.4531 & 0.4157 & 0.3148 \\
			Sidewalk & 0.5990 & 0.8502 & 0.5972 & 0.4974 & & Sidewalk & 0.5130 & 0.8904 & 0.4691 & 0.4211 \\
			Building & 0.5011 & 0.3995 & 0.8511 & 0.3019 & & Building & 0.4853 & 0.3154 & 0.9003 & 0.2117 \\
			Car & 0.4996 & 0.5018 & 0.5810 & 0.9051 & & Car & 0.3548 & 0.4183 & 0.4519 & 0.9331 \\
			\hline
		\end{tabular}
	}
\end{table*}

\begin{table*}[!t]
	\centering
	{
		\caption{The similarity of different classes between the learned class-wise features and the augmented domain prototypes. The first row represents the learned class-wise features and the first column represents the augmented domain prototypes.}
		\label{tab:dc4}
		\renewcommand\tabcolsep{2.0pt}
		\begin{tabular}{cccccc|ccccc}
			\hline
			& 
			\multicolumn{4}{c}{PCL}& 
			&
			&
			\multicolumn{4}{c}{CDPCL}
			\\
			\hline
			Class & Road & Sidewalk & Building & Car & & Class & Road & Sidewalk & Building & Car\\
			Road & 0.7998 & 0.6131 & 0.3377 & 0.4957 & & Road & 0.8512 & 0.4937 & 0.2903 & 0.3304 \\
			Sidewalk & 0.6982 & 0.7573 & 0.3984 & 0.5965 & & Sidewalk & 0.5701 & 0.8210 & 0.2965 & 0.3899 \\
			Building & 0.4005 & 0.3002 & 0.8007 & 0.3012 & & Building & 0.2418 & 0.2419 & 0.8411 & 0.1944 \\
			Car & 0.5190 & 0.4921 & 0.2910 & 0.8793 & & Car & 0.3941 & 0.3823 & 0.2740 & 0.8999 \\
			\hline
		\end{tabular}
	}
\end{table*}

\textbf{Quantitative Results.} We respectively use the cosine similarity and the Manhattan distance to measure the discrepancy changes between the learned class-wise features and the different domains before and after applying the proposed approach. The discrepancy changes of the ``road'', ``sidewalk'', ``building'', and ``car'' classes are shown in Table \ref{tab:dc1}, Table \ref{tab:dc2}, Table \ref{tab:dc3}, and Table \ref{tab:dc4}.

First, we compare the discrepancy measured by the Manhattan distance. The discrepancy between the learned class-wise features and the source domains is shown in Table \ref{tab:dc1}. The discrepancy between the learned class-wise features and the augmented domains is shown in Table \ref{tab:dc2}. From Table \ref{tab:dc1}, compared with the conventional PCL, the discrepancy between the learned class-wise features and the prototypes of the source domain is reduced in the classes of ``road'', ``building'', and ``car'' by using the proposed approach (CDPCL). From Table \ref{tab:dc2}, compared with the conventional PCL, the discrepancy between the learned class-wise features and the prototypes of the augmented domain is reduced in the classes of ``road'', ``sidewalk'', ``building'', and ``car'' by using the CDPCL. In addition, although the discrepancy of the ``sidewalk'' class between the learned class-wise features and the prototypes of the source domain is increased, the discrepancy of the ``sidewalk'' class between the learned class-wise features and the prototypes of the augmented domain is significantly decreased. We argue that this phenomenon is caused by the big gap between the ``sidewalk'' class prototypes of the source and augmented domains, which means the ``sidewalk'' class prototypes of the source domain are uncertain. Thus, a small weight is assigned to the ``sidewalk'' class prototypes of the source domain.

Second, we compare the discrepancy measured by the cosine similarity. The discrepancy between the learned class-wise features and the source domains is shown in Table \ref{tab:dc3}. The discrepancy between the learned class-wise features and the augmented domains is shown in Table \ref{tab:dc4}. From Table \ref{tab:dc3} and Table \ref{tab:dc4}, compared with the conventional PCL, the similarity of different classes is reduced while the similarity of the same class is significantly increased by using the proposed approach (CDPCL).

In conclusion, from these qualitative and quantitative experiments, the proposed approach can reduce the discrepancy between the learned class-wise features and the prototypes of different domains.

\section{Supplement Material}
\label{app_sm}
The supplement material will be released at https://github.com/se-\\abearlmx/CDPCL.

\end{document}